\documentclass[lettersize,journal]{IEEEtran}
\usepackage[colorlinks=true, linkcolor=blue, citecolor=blue, urlcolor=blue]{hyperref}

\usepackage{amsmath,amsfonts}
\usepackage{algorithmic}
\usepackage{algorithm}
\usepackage{array}
\usepackage[caption=false,font=normalsize,labelfont=sf,textfont=sf]{subfig}
\usepackage{textcomp}
\usepackage{stfloats}
\usepackage{url}
\usepackage{verbatim}
\usepackage{makecell}
\usepackage{nicematrix, xcolor}
\usepackage{multirow}
\usepackage{booktabs}
\usepackage{graphicx}
\usepackage{cite}
\usepackage{lineno}

\usepackage{listings}

\usepackage{cite}
\usepackage{manyfoot}

\usepackage{xcolor}
\usepackage{textcomp}
\usepackage{pifont}
\usepackage{soul}
\usepackage{tikz}

\usepackage[table]{xcolor}
\definecolor{customgreen}{HTML}{00B050}
\definecolor{orcidgreen}{HTML}{A6CE39}

\newcommand{\x}[1]{\textcolor{black}{#1}}
\newcommand{\bl}[1]{\textcolor{black}{#1}}
\newcommand{\orcidicon}[1]{%
  \textsuperscript{%
    \href{https://orcid.org/#1}{%
      \begin{tikzpicture}[baseline=-0.1em]
        \fill[orcidgreen] (0,0) circle (1.5ex);
        \node[white, scale=0.8, font=\bfseries\sffamily] at (0,0) {iD};
      \end{tikzpicture}%
    }%
  }%
}

\makeatletter
\let\NAT@parse\undefined
\makeatother

\usepackage[colorlinks=true, linkcolor=blue, citecolor=blue, urlcolor=blue]{hyperref}

\begin{document}
\title{BioAutoML-NAS: An End-to-End AutoML Framework for Multimodal Insect Classification via Neural Architecture Search on Large-Scale Biodiversity Data}
\author{Arefin Ittesafun Abian\orcidicon{0009-0003-4451-0838}, Debopom Sutradhar\orcidicon{0009-0008-3811-0228}, Md Rafi Ur Rashid\orcidicon{0000-0002-3089-5277},
Reem E. Mohamed\orcidicon{0000-0002-6992-6608}, Md Rafiqul Islam\orcidicon{0000-0001-7209-3881}, Asif Karim\orcidicon{0000-0001-8532-6816}, Kheng Cher Yeo \orcidicon{0000-0002-0453-3248}, Sami Azam\orcidicon{0000-0001-7572-9750}

\thanks{Manuscript Submitted 7\textsuperscript{th} October, 2025}

\thanks{(Corresponding authors: Arefin Ittesafun Abian and Sami Azam)}

\thanks{This work did not involve human subjects or animals in its research.}

\thanks{Arefin Ittesafun Abian, and Debopom Sutradhar are affiliated with the Department of Computer Science and Engineering, United International University, Dhaka, Bangladesh also with Applied Artificial Intelligence and Intelligent Systems (AAIINS) Laboratory, 1217, Dhaka, Bangladesh. (e-mail: aabian191042@bscse.uiu.ac.bd, dsutradhar201046@bscse.uiu.ac.bd).}
\thanks{Md Rafi Ur Rashid are affiliated with the Department of Computer Science and Engineering, Penn State University, University Park, PA, USA. (e-mail: mur5028@psu.edu).}

\thanks{Reem E. Mohamed is with the Faculty of Science and Information Technology, Charles Darwin University, Sydney, NSW, Australia (e-mail: reem.sherif@cdu.edu.au).}
\thanks{Md Rafiqul Islam, Asif Karim,  Kheng Cher Yeo and Sami Azam are with the Faculty of Science and Technology, Charles Darwin University, Casuarina, 0909, NT, Australia (e-mail: mdrafiqul.islam@cdu.edu.au, Charles.Yeo@cdu.edu.au, asif.karim@cdu.edu.au, Sami.Azam@cdu.edu.au).}}

\markboth{IEEE Transactions on Big Data,~Vol.~00, No.~0, August~2025}%
{Shell \MakeLowercase{\textit{et al.}}: A Sample Article Using IEEEtran.cls for IEEE Journals}


\maketitle

\begin{abstract}

Insect classification is important for agricultural management and ecological research, as it directly affects crop health and production. However, this task remains challenging due to the complex characteristics of insects, class imbalance, and large-scale datasets. To address these issues, we propose BioAutoML-NAS, the first BioAutoML model using multimodal data, including images, and metadata, which applies neural architecture search (NAS) for images to automatically learn the best operations for each connection within each cell. Multiple cells are stacked to form the full network, each extracting detailed image feature representations. A multimodal fusion module combines image embeddings with metadata, allowing the model to use both visual and categorical biological information to classify insects. An alternating bi-level optimization training strategy jointly updates network weights and architecture parameters, while zero operations remove less important connections, producing sparse, efficient, and high-performing architectures. Extensive evaluation on the BIOSCAN-5M dataset demonstrates that BioAutoML-NAS achieves 96.81\% accuracy, 97.46\% precision, 96.81\% recall, and a 97.05\% F1 score, outperforming state-of-the-art transfer learning, transformer, AutoML, and NAS methods by approximately 16\%, 10\%, and 8\% respectively. Further validation on the Insects-1M dataset obtains 93.25\% accuracy, 93.71\% precision, 92.74\% recall, and a 93.22\% F1 score. These results demonstrate that BioAutoML-NAS provides accurate, confident insect classification that supports modern sustainable farming.

\end{abstract}

\section{Introduction}
\IEEEPARstart{I}{nsect}
 classification is fundamental for the science of biodiversity and ecosystem management, as insects represent the most diverse and widespread biological community\cite{tan2024leveraging, dinca2025ensemble}. Precise classification enables distinguishing beneficial insects, such as pollinators and natural predators, from harmful pests, and with the rapid reproduction and evolution of pests, early detection is necessary to protect crops and natural habitats\cite{nguyen2024deep, Akhtar2025EdgeOptimized}. Several studies have relied on single-modal data, such as images, environmental metadata, or taxonomy \cite{truong2025insect}. In contrast, combining these sources through multimodal data provides a more comprehensive understanding of the situation from multiple perspectives. \cite{truong2025insect, tan2024leveraging}. However, multimodal data significantly increases the size of the dataset, creating challenges such as impractical manual processing, high insect diversity, and difficulty in collecting many species, leading to class imbalance\cite{longo2025improving, orsholm2025multi,dewi2024fruit}. Furthermore, conventional deep learning (DL) \cite{lecun2015deep}, transfer learning (TL) \cite{niu2021decade}, and transformer models with predefined architectures often fail to fully optimize performance on diverse data, making multimodal data essential and highlighting the value of approaches such as AutoML or Neural Architecture Search (NAS) \cite{ong2022next, li2019blockchain}.\par

 Recently, deep learning (DL) and machine learning (ML) models have made progress in utilizing large datasets and improving insect classification accuracy, yet key limitations persist. For large-scale datasets, models such as multi-axis vision transformer (MViT) \cite{pacal2024enhancing} have been applied to image classification, U-Net Ensemble \cite{wu2023deep} has used K-means post-processing for pixel-level segmentation, and Segment Anything in Images and Videos (SAM-2) \cite{ravi2024sam} has employed memory-guided attention for video object tracking and segmentation. However, high-capacity models often struggle to generalize due to reliance on hand-crafted post-processing and synthetic data, which limits spatial discrimination and real-world performance \cite{pacal2024enhancing, nguyen2024deep, wu2023deep, ravi2024sam}. 
 
In insect classification, various CNN and transfer learning (TL) models, including multilayer CNN \cite{balingbing2024application}, Deep Wide (DeWi) \cite{nguyen2024deep}, ViT \cite{dinca2025ensemble}, MobileViT, and EfficientNetV2B2 \cite{Akhtar2025EdgeOptimized}, have been applied; nevertheless, they often rely on manually designed CNNs with limited operation diversity, lacking modules necessary to capture complex morphological and contextual features \cite{Akhtar2025EdgeOptimized}. Most approaches focus solely on image data, overlooking valuable structured metadata such as species taxonomy or DNA barcoding. Specialized architectures have also been explored, including the Two-Branch Self-Correlation Network (TBSCN) \cite{tan2024leveraging}, which integrates principal component analysis (PCA) and spectral-spatial branches for hyperspectral imaging, and Swin-AARNet (Attention Augmented Residual Network) \cite{wang2025swin}, which employs depthwise weighted and global spatial attention for multi-scale feature extraction. Despite these advances, existing architectures remain rigid and resource-intensive, lacking adaptive efficiency mechanisms and sparsity. Additionally, many methods use fixed architectures throughout training, limiting the flexibility needed to co-optimize both structure and weights for improved performance \cite{dinca2025ensemble, wang2025swin, tan2024leveraging, Akhtar2025EdgeOptimized}.
\par

\begin{figure}[ht!]
\centering
\includegraphics[scale=0.22]{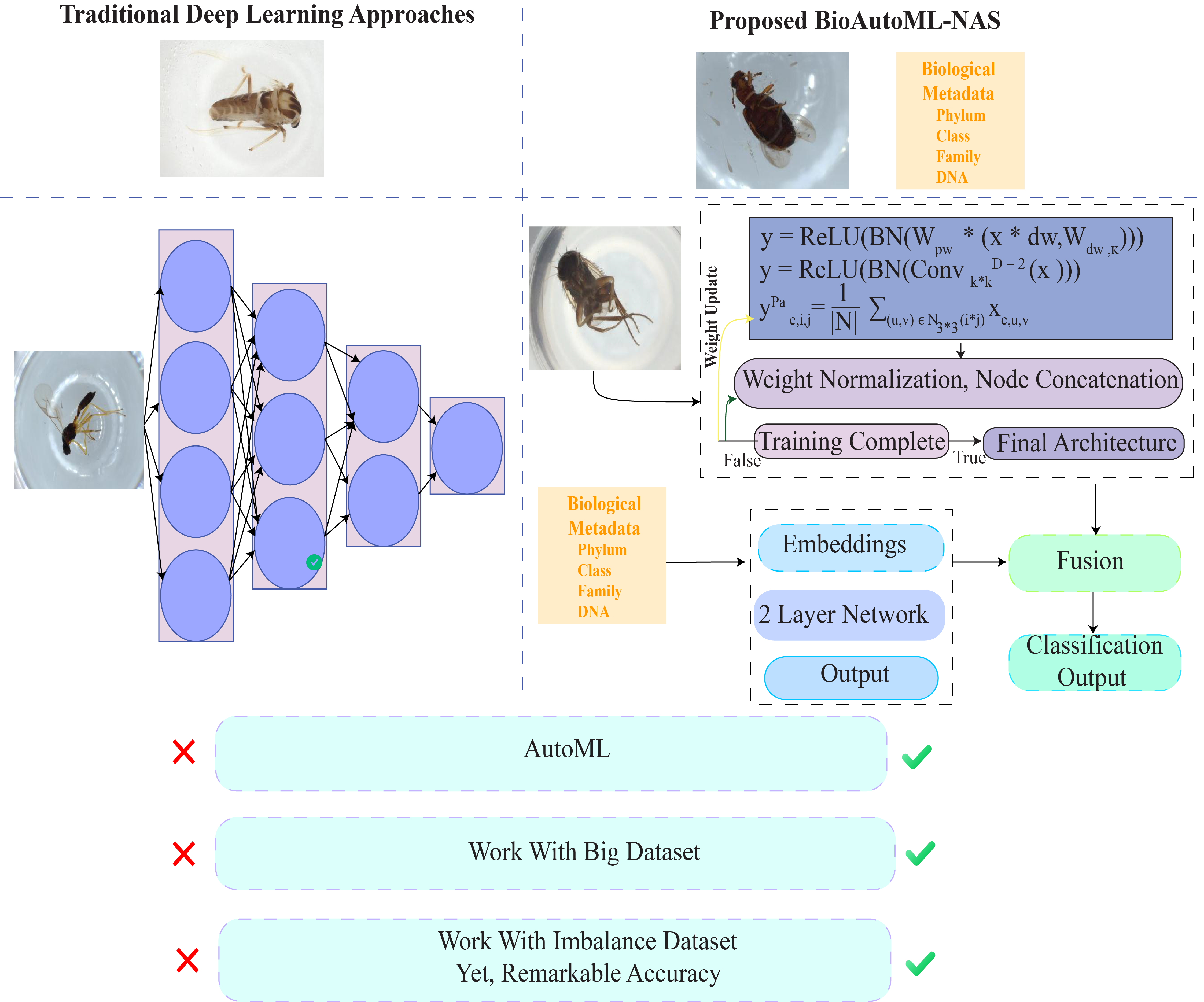} 
\caption{We process the images using a NAS-based image encoder and the metadata using a separate encoder, and fused the two representations to obtain the classification output.}
\label{fig:introdg}
\end{figure}

Driven by critical gaps in existing methods, we develop BioAutoML-NAS, an innovative end-to-end AutoML framework utilizing NAS. It has been designed to operate on the large-scale multimodal BIOSCAN-5M \cite{gharaee2024bioscan} biodiversity dataset, which includes both images, and metadata, and validated on the Insects-1M dataset \cite{nguyen2024insect}. The framework learns directly from biological data and derived compact, context-aware architectures that improved generalization without relying on synthetic data, data augmentation, or hand-crafted postprocessing. It employs gradient-based NAS within a diverse, biologically informed search space to automatically design optimal multi-scale feature extractors. Additionally, it integrates multimodal data through a dedicated Metadata Encoder and a fusion module to improve discriminative power. \bl{We introduce a dynamic, sparsity-driven NAS framework in which zero operations act as learnable pruning mechanisms to produce efficient and scalable architectures. Unlike static designs, the proposed approach continuously refines the network structure during training, enabling adaptive architecture evolution guided by data. This provides a distinct contribution to AutoML/NAS formulation, enabling data-driven, self-optimizing architecture learning instead of relying on fixed or generic designs. Although the final fusion is performed via concatenation, it follows modality-specific encoders that independently learn rich image and metadata representations within a shared latent space. This design enables interaction through aligned representations and joint learning in the embedding space rather than dedicated cross-modal fusion operations. Through this approach, our model successfully addressed three critical challenges: managing class imbalance, processing large-scale datasets, and introducing AutoML solutions specifically for insect classification, as presented in Figure \ref{fig:introdg}}. It attained an accuracy of 96.81\%, precision of 97.46\%, recall of 96.81\%, and an F1-score of 97.05\% , outperforming all existing methods in classifying insects. \par
The main contributions of this work are listed below. 
\begin{itemize}
    \item 
     To the best of our knowledge, BioAutoML-NAS is the first Bio AutoML model trained on multimodal data comprising images, and metadata for insect classification. This novel approach to ecological inference represents a significant advancement in large-scale, automated, and biologically informed model discovery.
    \item 
    BioAutoML-NAS integrates multimodal fusion, alternating bi-level optimization training, and zero operations to deliver sparse, high-performing architectures with reduced complexity and enhanced scalability.
    \item Our model mitigates data imbalance through label smoothing (0.1) and dropout-regularized multimodal fusion, enhancing generalization and inter-class separability. NAS-based optimization further ensures robust and balanced learning without re-sampling or class weighting.
    \item Despite the immense size and complexity of the dataset, BioAutoML-NAS achieves an accuracy of 96.81\%, outperforming all previously reported methods and establishing a new state-of-the-art (SOTA) in large-scale insect classification for biodiversity research.
    \item Our proposed model outperforms transformer-based and TL approaches, AutoML and NAS-based models, as well as other methods reported in the current literature,
    demonstrating superior robustness and generalizability across complex biological large data.
\end{itemize}

The organization of this article is as follows. Section \ref{LR} reviews related work in this domain. In Section \ref{method}, we describe the overall architecture of the proposed BIOSCAN-5M model. Sections \ref{experimental} and \ref{res} present the experimental setup and the performance of the model. Section \ref{discuss} discusses the significance of our findings in the context of current research and highlights potential directions for future work. Finally, Section \ref{conclusion} provides a summary of the study and concludes the article.

\section{Related Works}
\label{LR}
In this section, we review previous research across several domains, including NAS-driven models, large data-driven approaches, and insect classification. 
\subsection{NAS-Driven Model}
Recent advances \cite{saeedizadeh2024new, saeed20253d, dewi2024fruit, broni2024unsupervised, liang2025evolutionary} in NAS-driven models have focused on automating neural network design and evaluating their performance. Initially, Saeedizadeh et al. \cite{saeedizadeh2024new} used a gradient-based NAS framework to optimize cell design in a U-Net-style architecture with depths ranging from two to eight layers. 
\x{Here, the search space is constrained to a U-Net-style encoder–decoder with predefined reduction and expansion cells, limiting exploration to a fixed architectural pattern and reducing adaptability beyond segmentation tasks. In contrast, our framework eliminates such structural constraints and learns architectures directly from data without relying on predefined templates.}

Furthermore, Saeed et al. \cite{saeed20253d} proposed a 3D NAS architecture based on Point-Voxel Convolution (PVConv) as the core operator, employing a weight-sharing evolutionary search, \x{where the search space is limited to depth and channel width variations in a PVConv-based supernetwork, restricting it to a single building block. Random path sampling with weight sharing may also introduce optimization bias and weak structural guidance. In contrast, our method searches over diverse feature transformations within structured cells, enabling richer representation learning beyond simple scaling.}

In another study, Broni-Bediako et al. \cite{broni2024unsupervised} integrated Markov Random Field-based NAS with a self-training domain adaptation strategy, employing confidence- and energy-based pseudo-labeling to address cross-domain shifts, \x{where the search space is defined through factorized pairwise dependencies over discrete architectural choices, it is restricted to predefined kernel–label combinations, limiting complex interactions and multimodal learning. In contrast, our framework jointly learns architecture and multimodal representations from visual and metadata features.} Similarly, Liang et al. \cite{liang2025evolutionary} proposed an evolutionary NAS framework that constructs CNNs by assembling optimal modules, guided by diversity enhancement strategies and random forest-based fitness estimation. \x{Their search space uses a two-level binary encoding that separates network configuration and connectivity, resulting in a fragmented search process that prevents joint optimization of feature transformation and structure. In contrast, our framework uses a unified differentiable formulation, enabling co-adaptation of feature transformations and connectivity during training.}
\subsection{Lagre Data-Driven Approach}
Recent research \cite{zhang2024champ, pacal2024enhancing, wu2023deep, ravi2024sam} has explored the use of large-scale datasets and approaches to managing the challenges associated with their volume and complexity. 
Zhang et al. \cite{zhang2024champ} used Composite Motion Synthesis, Composite GCN, and a partition policy network for motion generation, prediction, and optimization. However, synthetic motions may limit modeling of real composite dynamics, reducing prediction consistency. Furthermore, Pacal et al. \cite{pacal2024enhancing} proposed MViT with SE blocks and a GRN-based MLP for a four-class dataset. However, its higher complexity may cause overfitting and reduce robustness to unseen data. Wu et al. \cite{wu2023deep} proposed a U-Net-based ensemble with skip connections for pixel-level segmentation, using K-fold cross-validation and K-means post-processing. However, K-means may limit separation of closely overlapping objects, affecting counting and localization accuracy. Finally, Ravi et al. \cite{ravi2024sam} proposed SAM-2, which extends SAM to videos by combining hierarchical encoding, memory-guided attention, and a lightweight mask decoder to track and segment objects despite occlusions. However, rapid appearance changes and severe occlusions reduce the precision of its segmentation.
\subsection{Insect Classification}
Recent studies \cite{balingbing2024application, tan2024leveraging, nguyen2024deep, dinca2025ensemble, wang2025swin, venkateswara2025deep, Akhtar2025EdgeOptimized}
have applied computational methods to the classification of insects. To begin with, B. Bilingual\cite{balingbing2024application} developed a multilayer CNN model, including two convolutional layers, max pooling, dropout layers, and dense layers, achieving an average classification accuracy of 84.51\%. Subsequently, Tan et al. \cite{tan2024leveraging} proposed TBSCN for hyperspectral images, integrating PCA-based dimensionality reduction with spectral and spatial patch correlation branches, achieving 93.96\% accuracy. Furthermore, Nguyen et al. \cite{nguyen2024deep} proposed DeWi, a CNN framework combining deep and wide steps with triplet margin and cross-entropy losses, achieving 76.44\% accuracy. In another study, Dinca et al. \cite{dinca2025ensemble} improved ViT fusion using a logistic regression metaclassifier on scaled, weighted logits, outperforming averaging and majority voting, and achieving 83.71\% accuracy. Furthermore, Wang et al. \cite{wang2025swin} proposed Swin-AARNet, a hierarchical model capturing multiscale context using depthwise weighted attention and global spatial attention, achieving 78.77\% accuracy. In another study, Akhtar et al. \cite{Akhtar2025EdgeOptimized} used MobileViT and EfficientNetV2B2 with post-training, quantization-aware, and representative data quantization, achieving 77.8\% accuracy. Finally, Venkateswara et al. \cite{venkateswara2025deep} proposed a CNN with dropout, batch normalization, and early stopping on segmented images, achieving 84.95\% accuracy.
\begin{figure*}[ht!]
\centering
\includegraphics[scale=0.32]{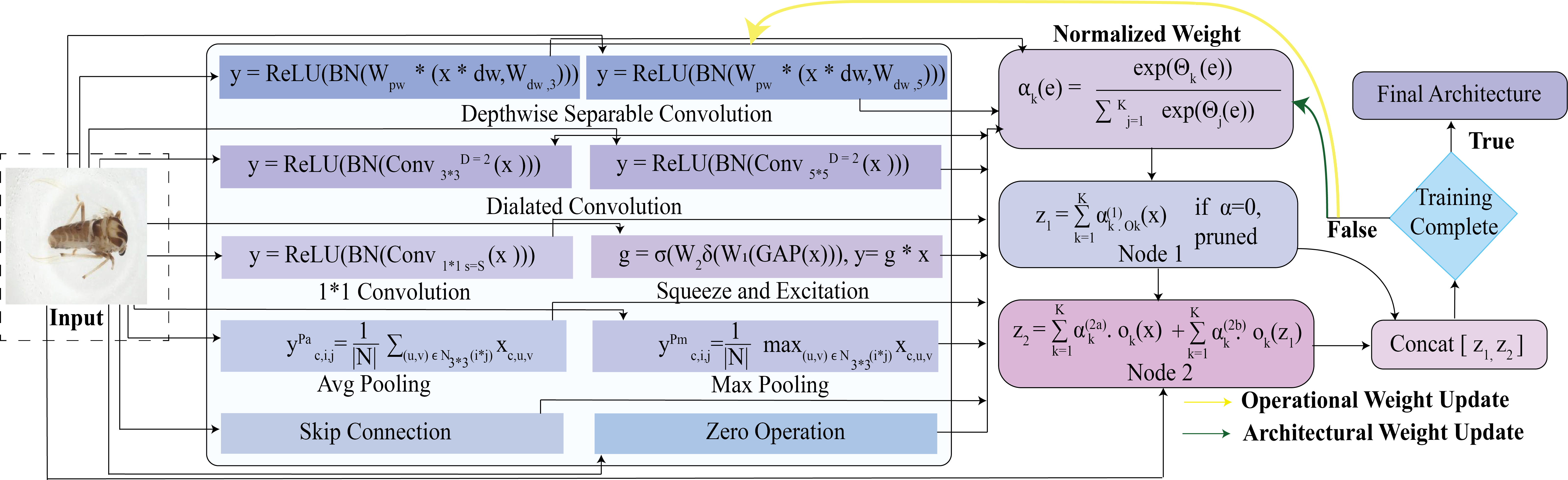} 
\caption{The NAS-based image encoder explores a search space of ten candidate operations, including convolutional filters, pooling layers, skip connections, and channel attention mechanisms, to automatically learn rich and detailed feature representations from input images.}
\label{fig:NAS}
\end{figure*}
 
Our proposed model, BioAutoML-NAS, overcomes critical limitations of existing approaches by introducing novel insights and solutions. Although traditional methods struggled with individual challenges, our model simultaneously tackles class imbalance, large-scale data processing, and the lack of AutoML solutions for insect classification. To our knowledge, no prior work has addressed all three challenges together using AutoML techniques. \x{Existing NAS search spaces are typically restricted to fixed templates, block-level scaling, or decoupled structural encodings, which limit coordinated modeling of feature transformations and connectivity. Unlike these approaches, our search space defines ten heterogeneous primitive operations within a structured cell, covering diverse convolutional, pooling, attention, and identity-based transformations for rich feature extraction. This enables combined exploration of operation selection and architectural composition, leading to more expressive representations for multimodal classification.} We use the BIOSCAN-5M biodiversity dataset, which includes images, and metadata, to develop the first NAS-based multimodal AutoML model, representing a previously unexplored direction in environmental research. \bl{Furthermore, the proposed method introduces a task-specific, gradient-based NAS framework with a biologically informed, sparsity-aware search space and structured dual-node cells for multi-scale feature learning and controlled feature interaction. It integrates multimodal information through a metadata encoder and fusion module, while zero operations and dynamic training enable efficient, scalable, and sparse architecture construction. Through strict alternating bi-level optimization, the model jointly learns architecture and weights in a stable manner, representing a meaningful contribution to AutoML/NAS for multimodal biological classification.}

\section{Methodology}
\label{method}
In this study, we adopt a systematic methodology that integrates data preprocessing, model design, training, and evaluation to ensure robustness and reproducibility. The approach emphasizes handling class imbalance, applying NAS for optimized feature extraction, and incorporating multimodal integration strategies. 

\subsection{Proposed Model: BioAutoML-NAS}
We develop the BioAutoML-NAS model, which comprises two primary encoders: an image encoder and a metadata encoder. The image encoder employs NAS to automatically identify an optimal architecture from a predefined set of ten primitive operations, facilitating the extraction of detailed and context-aware visual representations. Simultaneously, the metadata encoder encodes biological descriptors, including DNA barcodes, hierarchical taxonomic ranks, and order-level labels, into dense feature embeddings. A fusion module subsequently integrates the visual and metadata representations into a unified feature space for robust classification.
\subsubsection{Neural Architecture Search for Image Encoding}
In our proposed model, the image encoder search space has been designed to balance representational diversity and computational efficiency for biological image analysis.  It comprises ten primitive operations, each serving a distinct functional role in feature extraction, contextual modeling, and information flow, as shown in Figure \ref{fig:NAS}.

The first two operations are depthwise separable convolutions, denoted as D3 and D5, corresponding to kernel sizes $3\times3$ and $5 \times 5$, respectively \cite{lu2021optimizing}. Each operation applies a depthwise convolution using the filter $W_{dw,k}$ (with $k=3$ for D3 and $k=5$ for D5), followed by a $1 \times 1$ pointwise convolution with filter $W_{pw}$, batch normalization (BN) to stabilize and accelerate training, and ReLU activation to introduce nonlinearity, as shown in Equation~\eqref{eq:depthwise_pointwise}:

\begin{equation}
y = \text{ReLU} \Big( \text{BN} \big( W_{pw} * (x *_{\text{dw}} W_{dw,k}) \big) \Big)
\label{eq:depthwise_pointwise}
\end{equation}

Here, $x$ denotes the input tensor and $*_{\text{dw}}$ indicates the depthwise convolution operation. The third and fourth operations are dilated convolutions, denoted $L_3$ and $L_5$, with kernel sizes $k=3$ for $L_3$ and $k=5$ for $L_5$, and a dilation rate of $D=2$ \cite{hu2025ddconv}, as defined in Equation (\ref{eq:dil_conv}):

\begin{equation}
\label{eq:dil_conv}
y = \text{ReLU}\Big(\text{BN}\big(\text{Conv}_{k\times k}^{D=2}(x)\big)\Big)
\end{equation}
Furthermore, we have a $1\times1$ convolution \cite{pang2017convolution}, denoted as $S$, which enables channel mixing and optional downsampling , as calculated in Equation (\ref{eq:S_conv}):
\begin{equation}
\label{eq:S_conv}
y = \text{ReLU} \Big( \text{BN} \big( \text{Conv}_{1\times1,\,s=S}(x) \big) \Big)
\end{equation}
Subsequently, we incorporated the Squeeze-and-Excitation (SE) \cite{shu2022expansion} block, denoted ${SE}$, to enhance channel-level feature discrimination by amplifying informative signals and suppressing noise, calculated using Equation (\ref{eq:SE}):
\begin{equation}
\label{eq:SE}
    g = \sigma \left( W_2 \, \delta \left( W_1 \, \mathrm{GAP}(x) \right) \right), \quad
    y = x \odot g
\end{equation}
where $x$ is the input tensor with $C$ channels, $\mathrm{GAP}$ is the global average pooling, $W_1 \in \mathbb{R}^{\frac{C}{r} \times C}$ and $W_2 \in \mathbb{R}^{C\times \frac{C}{r}}$ are learnable weight matrices of two fully connected layers, $r$ is the reduction ratio controlling the bottleneck, $\delta(\cdot)$ is the activation of ReLU \cite{wang2019learning}, $\sigma(\cdot)$ is the sigmoid function, $g$ is the learned channel weight vector, and $\odot$ denotes element-wise multiplication. 
The seventh and eighth operations are local pooling, denoted as $P_a$ (average pooling) \cite{sun2021ampnet} and $P_m$ (max pooling) \cite{sun2021ampnet}, which summarize $3\times3$ spatial neighborhoods to reduce noise and highlight salient features. These operations are computed in Equation (\ref{eq:pooling}):
\begin{equation}
\label{eq:pooling}
\begin{split}
y^{Pa}_{c,i,j} &= \frac{1}{|N|} \sum_{(u,v) \in N_{3\times3}(i,j)} x_{c,u,v}, \\
y^{Pm}_{c,i,j} &= \max_{(u,v) \in N_{3\times3}(i,j)} x_{c,u,v}
\end{split}
\end{equation}

where $x$ is the input tensor, $c$ indexes the channel, $(i,j)$ indexes the spatial position, $(u,v)$ iterates over the $3\times3$ neighborhood $N_{3\times3}(i,j)$ around $(i,j)$, $|N|$ is the number of elements in the neighborhood ($|N|=9$), $y^{Pa}$ denotes the average-pooled output, and $y^{Pm}$ denotes the max-pooled output. 
Finally, skip connections \cite{zhou2019unet++} and zero operations, denoted $SK$ and $Z$, serve as key mechanisms for shortcutting and pruning. For stride $s=1$, skip connections directly forward the input ($y^sK = x$), while $Z$ produces an all-zero tensor ($y^Z = 0$). When $s>1$, skip connections down-sampling and concatenate shifted features, while $Z$ outputs a spatially downsampled zero tensor, encouraging architectural sparsity and controlled information flow.

To transform discrete operation selection into a continuous and learnable form, each directed edge \(e\) in the computational cell is expressed as a weighted combination of all candidate operations. Each edge is parameterized by a set of learnable weights \(\theta^{(e)}=\{\theta_k^{(e)}\}_{k=1}^{K}\), where \(K\) is the number of candidate operations. During the forward pass, the relative contribution of each primitive operation \(o_k(\cdot)\) is determined using a softmax transformation. Formally, the normalized selection weight is computed as using Equation (\ref{eq:softmax}):
\begin{equation}
\alpha_k^{(e)} \;=\; \frac{\exp\big(\theta_k^{(e)}\big)}{\displaystyle\sum_{j=1}^{K} \exp\big(\theta_j^{(e)}\big)} \label{eq:softmax}
\end{equation}
Here \(\alpha_k^{(e)}\) denotes the normalized weight of the \(k\)-th candidate operation on edge \(e\). The softmax ensures \(\sum_{k=1}^K \alpha_k^{(e)} = 1\), enabling differentiable selection among operations during architecture search \cite{chen2022approximate}. Each operation contributes according to its learned weight, allowing the network to prioritize the most relevant transformations. During training, less important operations are gradually suppressed while more significant ones dominate. Upon convergence, the final discrete architecture is obtained by selecting, for each node, the operation with the highest \(\alpha\).
\begin{figure*}[ht!]
\centering
\includegraphics[scale=0.3]{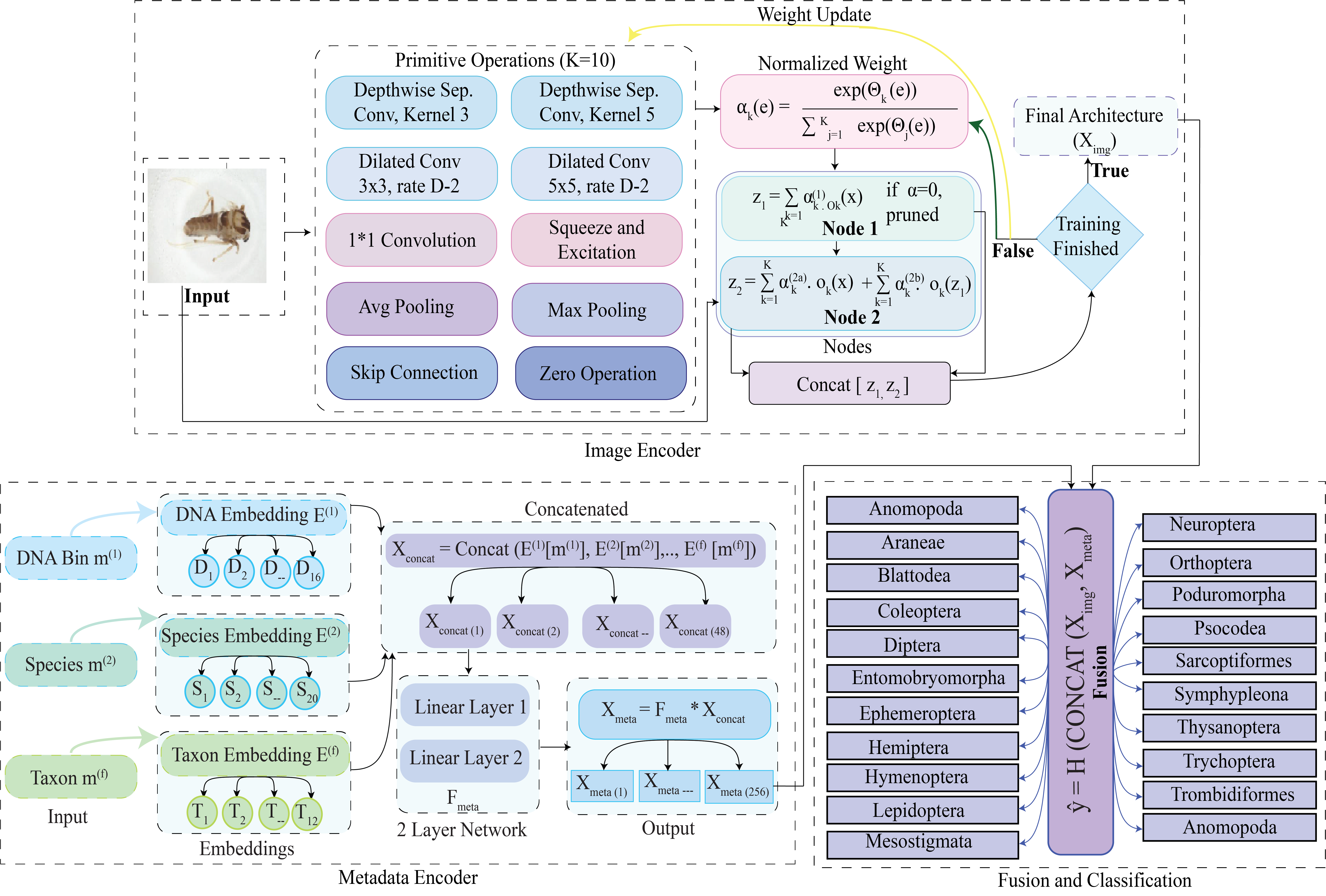} 
\caption{Overview of the proposed BioAutoML-NAS model, highlighting dual encoders that extract multimodal features and a fusion module that integrates them for accurate classification.}
\label{fig:maindg}
\end{figure*}

At the core of the model is a modular searchable cell, the fundamental building block of the network. Each cell contains two intermediate computational nodes \cite{guo2019blockchain}. The output of edge \(e\) is computed as the weighted sum of all candidate operations. Node 1 receives input feature map \(x\in\mathbb{R}^{C\times H\times W}\) and applies all \(K\) candidate operations in parallel; the outputs are aggregated via a softmax-weighted sum, as defined it Equation (\ref{eq:z1}):
\begin{equation}
z_1 \;=\; \sum_{k=1}^K \alpha_k^{(1)}\, o_k(x), \label{eq:z1}
\end{equation}
Here, \(o_k(\cdot)\) denotes the \(k\)-th primitive operation, and the normalized weight \(\alpha_k^{(1)}\) for node 1 is obtained through a softmax function.

The second node combines the original input \(x\) and the transformed output \(z_1\) from the first node. Each of these two inputs is processed independently by the same set of \(K\) candidate operations, each with its own learnable weights. The output of the second node is computed as defined in Equation (\ref{eq:z2}):
\begin{equation}
z_2 \;=\; \sum_{k=1}^K \alpha_k^{(2a)}\, o_k(x) \;+\; \sum_{k=1}^K \alpha_k^{(2b)}\, o_k(z_1), \label{eq:z2}
\end{equation}
where \(\alpha_k^{(2a)}\) and \(\alpha_k^{(2b)}\) correspond to the operation weights for the paths from \(x\) and \(z_1\), respectively.

Finally, the intermediate representations \(z_1\in\mathbb{R}^{C\times H\times W}\) and \(z_2\in\mathbb{R}^{C\times H\times W}\) are concatenated along the channel axis to form a joint feature map, according to Equation (\ref{eq:concat}):
\begin{equation}
\tilde{z} \;=\; \mathrm{Concat}\big[z_1, z_2\big], \qquad \tilde{z}\in\mathbb{R}^{2C\times H\times W}. \label{eq:concat}
\end{equation}

This concatenation preserves the distinct information pathways learned by each node while enabling richer cross-feature interactions. A point-wise 1$\times$1 convolution restores channel dimensionality, integrates features into a unified representation, reduces dimensionality back to \(C\), and performs channel mixing for downstream processing. To avoid unnecessary computation, operations with selection weights below a threshold (e.g., 10$^{-6}$) are omitted during the forward pass. This pruning discards less useful operations as training progresses, making the model more efficient and interpretable. The resulting features are subsequently merged with metadata in the multimodal integration framework.

\subsubsection{Multimodal Integration Framework}

To jointly adapt visual features and complementary biological metadata, we propose a dual-branch design consisting of an image encoder, a metadata encoder, and a fusion module to fuse the both encoders, as illustrated in Figure \ref{fig:maindg}.

The metadata encoder transforms discrete biological descriptors into a continuous latent representation that can be directly integrated with image features. Each categorical field $m^{(f)}$ (DNA barcoding bin, orders-level label, hierarchical taxonomic rank) is first embedded into a fixed-dimensional vector space. These embeddings are then concatenated to form a unified metadata vector, which is projected into the same dimensionality as the image feature vector to facilitate effective multimodal fusion.

Formally, the metadata representation is computed as Equation (\ref{eq:meta_enc}):
\begin{equation}
\label{eq:meta_enc}
\resizebox{\columnwidth}{!}{
\ensuremath{X_{\mathrm{meta}} = \mathcal{F}_{\mathrm{meta}}\!\left( \mathrm{Concat}\!\left[ E^{(1)}[m^{(1)}], E^{(2)}[m^{(2)}], \dots, E^{(F)}[m^{(F)}] \right] \right)}
}
\end{equation}
where $m^{(f)}$ is the $f$-th categorical metadata field, $E^{(f)} \in \mathbb{R}^{V_f \times d_f}$ is the learnable embedding matrix for field $f$ with vocabulary size $V_f$ and embedding dimension $d_f$, $E^{(f)}[m^{(f)}] \in \mathbb{R}^{d_f}$ is the embedded vector, $\mathrm{Concat}[\cdot]$ denotes concatenation across all fields, and $\mathcal{F}_{\mathrm{meta}}(\cdot)$ is a two-layer feedforward network projecting the concatenated vector into $X_{\mathrm{meta}} \in \mathbb{R}^{256}$.

The fusion module integrates the modality-specific representations and produces the final prediction by jointly reasoning over visual and metadata-derived cues. The final output is computed according to Equation (\ref{eq:output}):
\begin{equation}
\label{eq:output}
\hat{y} = \mathcal{H}\!\left( \mathrm{Concat}\!\left[ X_{\mathrm{img}}, X_{\mathrm{meta}} \right] \right),
\end{equation}
where $X_{\mathrm{img}} \in \mathbb{R}^{256}$ is the image feature vector, $\mathcal{H}(\cdot)$ denotes the classification head, and $\hat{y}$ is the predicted probability distribution over the target classes.

\subsection{Training and Architecture Search Strategy}

\begin{figure}[ht!]
\centering
\includegraphics[scale=0.33]{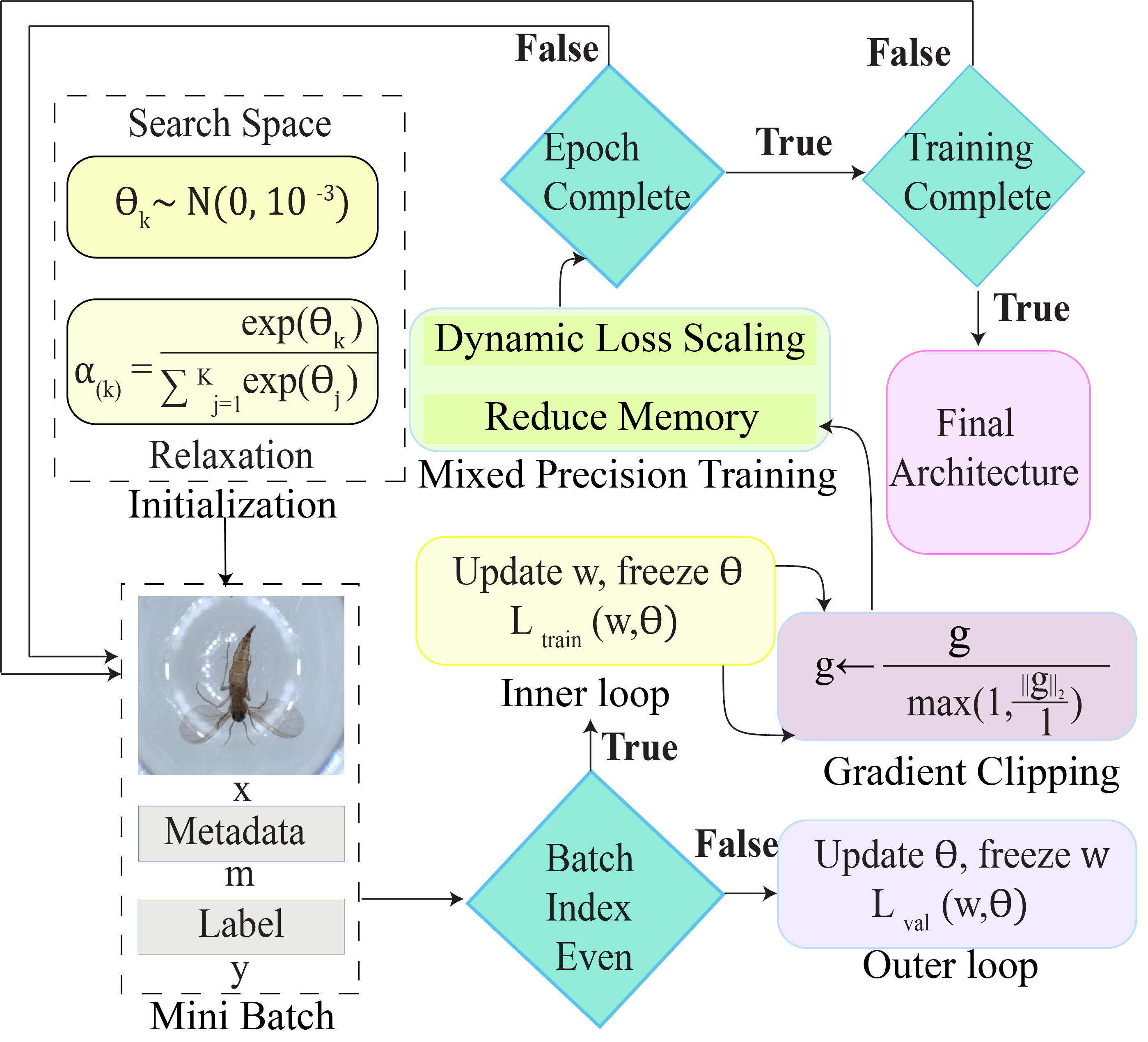} 
\caption{In the bi-level training framework, architectural parameters are updated on odd-numbered batches, while network weights are updated on even-numbered batches, enabling stable optimization by alternating between architecture parameter updates and weight learning.}
\label{fig:training}
\end{figure}

We have built the training framework for BioAutoML-NAS to jointly optimize network weights and architecture in a stable and computationally efficient manner \cite{ye2020distributed}. This is achieved through a differentiable NAS approach, where the search space is continuously relaxed and discretized at the end of training. Optimization follows a bi-level formulation, alternating between learning model parameters and selecting operations within each computational cell. This strategy allows the architecture to dynamically adapt to the dataset while preventing interference between weight updates and architectural decisions, as illustrated in Figure \ref{fig:training}.

\subsubsection{Alternating Bi-Level Optimization}

After defining the search space and multimodal feature extraction pipeline, the learning process jointly optimizes the network parameters and the architecture parameters that determine the operation selection at each edge \cite{liu2021investigating}. This is formulated as a bilevel optimization problem, where the inner objective updates model weights for a fixed architecture, and the outer objective refines the architecture based on validation performance. The architecture search is formalized using Equations
(\ref{eq:bilevel_outer}) and (\ref{eq:bilevel_inner}):
\begin{align}
\min_{\theta} \quad & \mathcal{L}_{\mathrm{val}}\!\left( w^*(\theta), \theta \right), \label{eq:bilevel_outer} \\
\quad w^*(\theta) &= \arg \min_{w} \ \mathcal{L}_{\mathrm{train}}(w, \theta), \label{eq:bilevel_inner}
\end{align}

where $w$ denotes the standard network parameters (convolutional filters, normalization layers, etc.), and $\theta$ represents the architecture parameters assigned to the candidate operations within the computational cell.

To solve this formulation efficiently, an alternating update scheme is applied at the mini-batch level. In even numbered mini-batches, $\theta$ is kept fixed and $w$ is updated using AdamW (learning rate $1\times10^{-3}$, weight decay $1\times10^{-4}$), focusing solely on representation learning. On odd-numbered mini-batches, $w$ is frozen and $\theta$ is updated using Adam (learning rate $3\times10^{-4}$, weight decay $1\times10^{-3}$), adapting the network topology to the learned representations. This separation mitigates gradient interference between the two sets of parameters, leading to a more stable search dynamics.

Each architecture parameter $\theta_k$ is initialized from a Gaussian distribution $\mathcal{N}(0, 10^{-3})$ and transformed into a normalized selection weight for its corresponding primitive operation using Equation (\ref{eq:bi}):
\begin{equation}
\label{eq:bi}
\alpha_k = \frac{\exp(\theta_k)}{\sum_{j=1}^K \exp(\theta_j)},
\end{equation}
where $\alpha_k$ is the probability of selecting the $k$-th operation, $K$ is the number of candidates on the edge, and $\sum_{k=1}^K \alpha_k = 1$. This continuous relaxation allows the search to softly explore all operations during the early stages before converging to the most probable choices.

To ensure numerical stability and efficient hardware utilization, all updates employ gradient clipping with a maximum norm of $1.0$ and mixed-precision training with dynamic loss scaling. These criteria reduce memory consumption, accelerate convergence without sacrificing accuracy, and ensure reproducibility. The complete optimization process, integrating both parameter and architecture updates, is described in Algorithm~\ref{alg:nas-training}.

\begin{algorithm}[!t]
\caption{Gradient-Based NAS Training with Alternating Optimization}
\label{alg:nas-training}
\begin{algorithmic}[1]
\REQUIRE Training data $\mathcal{D}$, initial parameters $\theta$ (architecture), $w$ (network)
\STATE Initialize $\theta \sim \mathcal{N}(0, 10^{-3})$, $w \sim \text{Initialization}$
\FOR{each epoch $e = 1$ to $E$}
    \FOR{each mini-batch $(x, m, y) \in \mathcal{D}$}
        \IF{batch index is even}
            \STATE Freeze $\theta$, unfreeze $w$
            \STATE Forward pass: compute loss $\mathcal{L}_{\text{task}}$
            \STATE Backpropagate and update $w$ using AdamW
        \ELSE
            \STATE Freeze $w$, unfreeze $\theta$
            \STATE Forward pass: compute loss $\mathcal{L}_{\text{task}}$
            \STATE Backpropagate and update $\theta$ using Adam
        \ENDIF
        \STATE Apply gradient clipping 
    \ENDFOR
\ENDFOR
\end{algorithmic}
\end{algorithm}

\subsubsection{Evaluation and System-Level Optimizations}

After training, the final architecture is obtained by selecting the operation with the highest probability on each edge, thereby converting the continuous search formulation into a fixed, deployable model. This deterministic configuration can be reproduced for evaluation, deployment, or transfer learning.

To support stable and efficient training during architecture search, the implementation integrates several key optimizations \cite{liu2021survey}. Mixed-precision training with NVIDIA AMP executes selected operations in FP16, reducing memory footprint and improving throughput, while gradient scaling ensures numerical stability. An adaptive batch size mechanism automatically selects the largest feasible batch (32–512) based on available GPU memory, maximizing utilization without manual tuning. A checkpointing system records model weights, architecture parameters, optimizer states, and training history at each epoch, enabling both recovery and retrospective analysis. Memory management is reinforced through explicit CUDA cache clearing and garbage collection, mitigating fragmentation and out-of-memory risks. Collectively, these measures establish a reproducible and hardware-efficient training environment that supports reliable exploration of complex architecture spaces.

\bl{The proposed method defines a task-specific NAS formulation designed for biologically complex, large and imbalanced data, moving beyond generic search spaces. The search space incorporates sparsity-driven architecture learning, where zero and skip operations function as learnable pruning mechanisms that progressively simplify the model during training. A structured dual-node cell design is introduced to regulate feature interaction and reduce the unstructured complexity commonly observed in standard NAS cells. The architecture is jointly optimized with a metadata encoder to enable aligned multimodal representation learning within a shared latent space, while a strict alternating bi-level optimization strategy stabilizes the coupled learning of network weights and architecture parameters. The final fusion is performed via concatenation; however, this follows independent encoding of image and metadata streams, where each encoder learns task-specific representations in a shared embedding space. This design preserves modality-specific richness while ensuring alignment at the feature level, enabling interaction through representation consistency rather than complex fusion operations. These design choices establish a unified AutoML/NAS framework tailored for multimodal biological classification under challenging data conditions.}

\section{Experimental Details}
\label{experimental}
This section describes the experimental setup for evaluating the proposed BioAutoML-NAS framework. We first detail the datasets and preprocessing procedures, followed by the training environment implementation.
\subsection{Data Preparation}

\subsubsection{Datasets}
In the present study, two separate public datasets are utilized to assess the efficiency of the classification model. Among the two datasets, the BIOSCAN-5M \cite{gharaee2024bioscan} dataset is used for training, validation, and testing, while the Insects-1M \cite{nguyen2024insect} dataset is used for cross-dataset validation.

\subsubsection{BIOSCAN-5M}
The BIOSCAN-5M dataset contains specimen of 5 million images at an original resolution of 1024×768 pixels, each retrievable through the processid field in the metadata. Further cropping and resizing are applied to reduce the overall image dimensions. In addition to images, it provides genetic data comprising raw nucleotide sequences (dna\_barcode) and Barcode Index Numbers (dna\_bin). Each record is uniquely identified by processid. Taxonomic labels are available across seven hierarchical ranks, denoted by the fields phylum, class, order, family, subfamily, genus, and species.

\subsubsection{Insects-1M}
The Insects-1M dataset contains 1,017,036 images covering 34,212 species with a detailed taxonomic hierarchy. It includes 15 classes and 91 orders, further organized into 54 suborders, 209 superfamilies, 1,189 families, 1,059 subfamilies, 1,315 tribes, 213 subtribes, and 11,127 genera, providing comprehensive multi-level taxonomic coverage of insect diversity.

\subsection{Data Preprocessing}
\x{For the BIOSCAN-5M training dataset, insect classification is performed at the order level. Although the order attribute shows class imbalance, the remaining taxonomic attributes present considerably greater challenges due to increased imbalance, data sparsity, and the large number of classes. Specifically, the class attribute is strongly imbalanced, with a single dominant class (Insecta) accounting for approximately 98\% of the samples. The family attribute consists of 934 classes, which introduces substantial complexity and makes reliable classification challenging. Furthermore, the subfamily, genus, and species attributes contain significant proportions of missing data, approximately 71\%, 76\%, and 91\%, respectively. In addition, these attributes also involve a very large number of classes, namely 1,542 for subfamily, 7,605 for genus, and 22,622 for species. Considering these limitations, classification at the order level is adopted as a more reliable and practically appropriate approach.} As the order attribute is also highly imbalanced, any order with fewer than 500 instances is grouped into an Other category. After this consolidation, the order includes 21 classes. In the training set, Diptera has the highest number of instances with 2,573,047 images, while Anomopoda has the fewest with only 164 images. \x{The BIOSCAN-5M dataset is partitioned into training, validation, and test sets at ratios of 80\%, 10\%, and 10\% respectively, using stratified random sampling at the specimen level to maintain consistent class distribution across all three splits.} 

We applied the same processing to our Insects-1M validation dataset, grouping any order with fewer than 500 instances into an Other category. After this adjustment, the dataset includes a total of 39 orders. Across the dataset, the largest number of instances is found in the Spiders (Araneae) order, with 87,836 images. The dataset remains highly imbalanced, with some orders containing very few instances; for example, Opilioacarida has only six images.

\subsection{\x{Implementation Setup}}
\x{All computational experiments are executed on a workstation configured with an AMD Ryzen 5 5600X six-core central processing unit (CPU) and 16 GB of system memory. Graphical and parallel processing tasks are supported by a ZOTAC GAMING GeForce RTX 3060 Twin Edge OC graphics processing unit (GPU) equipped with 12 GB of video memory (VRAM). The development environment employed is PyCharm 2025.2, with the implementation carried out in Python 3.13.2. GPU-accelerated computations utilize the NVIDIA CUDA Toolkit 12.8.0 in combination with cuDNN 9.10.2.}

\section{Results}
\label{res}
To evaluate the classification performance of the proposed BioAutoML-NAS framework, we conduct comprehensive experiments on the BIOSCAN-5M dataset and additionally evaluate generalization on the Insects-1M dataset. We present the results of our model and compare them with SOTA methods and existing literature.

\subsection{Performance of the proposed model}

\subsubsection{Evaluation metrics}
Our proposed model BioAutoML-NAS achieves an accuracy of 96 81\%, demonstrating its strong capability to accurately s  classify insects. In addition, it obtains a precision of 97.46\%, a recall of 96. 81\%, and an F1 score of 97. 05\%, providing further evidence of its robustness and reliability as a classification model.

To address severe class imbalance in the dataset, we further report macro-averaged metrics in Table~\ref{tab:eval_metrics}. The model achieves a Macro-F1 score of 80.29\%, Macro Precision of 80.17\%, and Macro Recall of 82.91\%. These metrics provide a class-balanced evaluation and confirm that the model maintains reasonable performance across both majority and minority classes. 

\begin{table}[!ht]
\centering
\caption{Comprehensive Evaluation Metrics of BioAutoML-NAS}
\label{tab:eval_metrics}
\begin{tabular}{lc}
\toprule
\textbf{Metric} & \textbf{Value (\%)} \\
\midrule
Accuracy & 96.81 \\
Precision (Weighted) & 97.46 \\
Recall (Weighted) & 96.81 \\
F1-Score (Weighted) & 97.05 \\
Precision (Macro) & 80.17 \\
Recall (Macro) & 82.91 \\
F1-Score (Macro) & 80.29 \\
Balanced Accuracy & 82.91 \\
\bottomrule
\end{tabular}
\end{table}

\bl{However, throughout the rest of the paper, the weighted F1, precision and recall are reported for all experiments, computed proportionally to class support. Weighted metrics are adopted for all comparative experiments as they reflect overall classification performance in a manner consistent with the class distribution of the dataset, and ensure a fair and standardized comparison across all baseline methods. The macro-averaged metrics reported in Table~\ref{tab:eval_metrics} are presented exclusively to provide a complementary. The macro metrics are presented only to show the model's robustness to imbalanced data.}

\subsubsection{Curve Analysis}

\begin{figure*}
    \centering
    \includegraphics[scale=0.2]{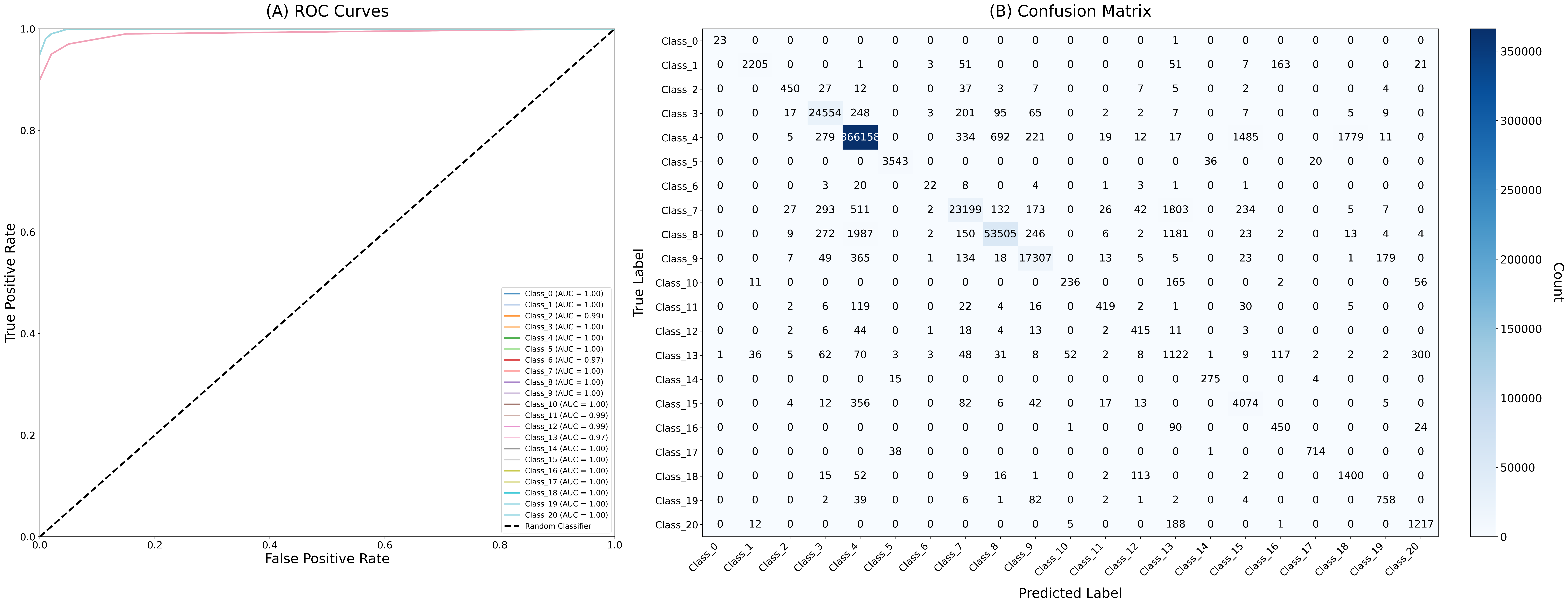}
    \caption{ROC curve and confusion matrix of the proposed BioAutoML-NAS model on the BIOSCAN-5M dataset, demonstrating its high classification accuracy and reliable performance across classes.}
    \label{Rocandconf}
\end{figure*}

The confusion matrix of our proposed model is presented in Figure \ref{Rocandconf}. The diagonal elements represent true positive cases, indicating correctly classified instances. Only a small number of misclassifications are observed. For instance, 36 instances of class 13 are misclassified as class 1, while 356 instances of class 15 are misclassified as class 4. Similarly, 51 instances of class 7 are predicted as class 1, and 27 instances are misclassified as class 2. Despite these minor errors, most samples are correctly classified, resulting in an overall accuracy of 96.81\%.

Figure \ref{Rocandconf} illustrates the Receiver Operating Characteristic (ROC) curves of the proposed BioAutoML-NAS model. Almost all classes achieve an AUC of 1.00, with only a few, such as Class 6 (0.97), Class 11 (0.99), Class 12 (0.99), and Class 13 (0.97), showing slightly lower but still high values. The curves remain close to the top-left corner, indicating high confidence and very low misclassification. Overall, the results demonstrate that BioAutoML-NAS effectively distinguishes insect species, even under class imbalance and high inter-species similarity.

\begin{figure*}
    \centering
    \includegraphics[scale=0.5]{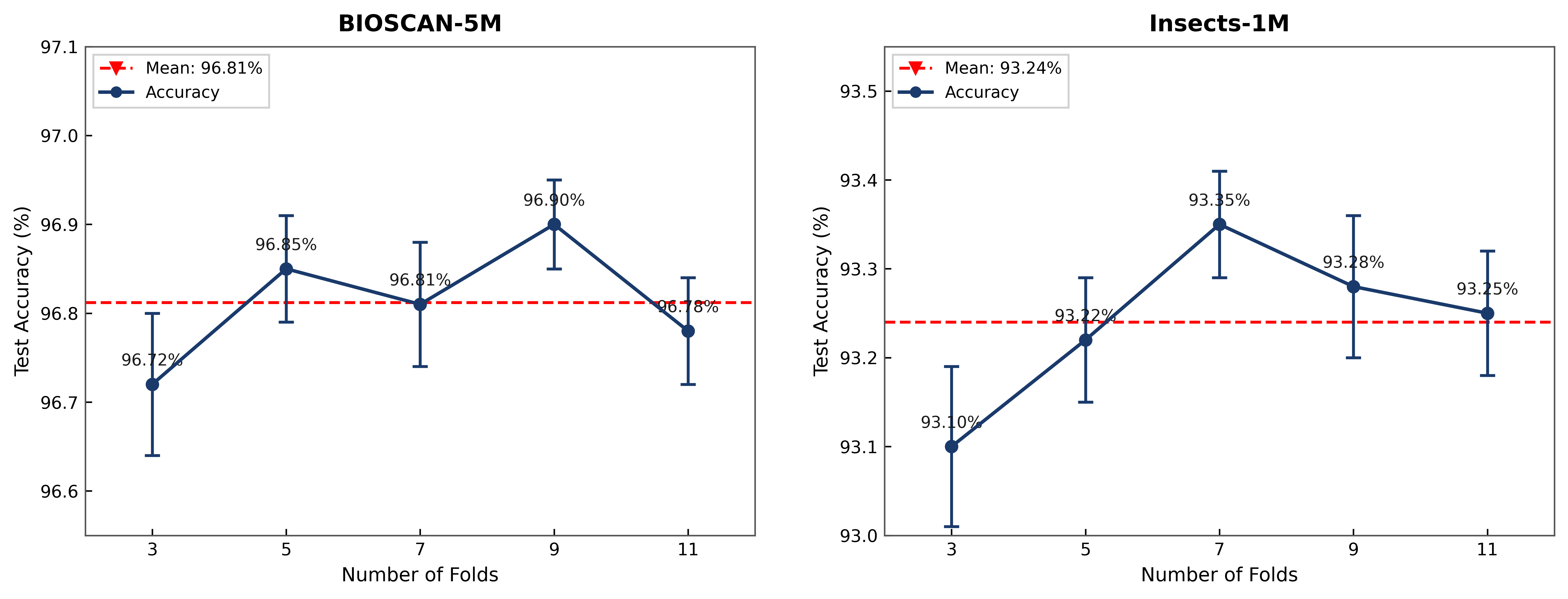}
    \caption{\x{K-fold cross-validation results for BioAutoML-NAS on BIOSCAN-5M and Insects-1M.}}
    \label{fig: kfold}
\end{figure*}

\x{To mitigate potential data leakage and ensure the validity of the evaluation, we employed Stratified Group K-fold Cross-Validation. Figure \ref{fig: kfold} presents the K-fold cross-validation results for BioAutoML-NAS on the BIOSCAN-5M and Insects-1M datasets. For BIOSCAN-5M, the accuracy fluctuates slightly across different folds, with a mean accuracy of 96.81\%. Similarly, for Insects-1M, the model shows consistent performance across folds, with a mean accuracy of 93.25\%. The error bars represent the standard deviation, demonstrating the stability and robustness of the model's performance across the different folds.}

\subsection{Ablation Study}

The ablation results in Table \ref{tab:ablation} clearly demonstrate how architectural choices influence both accuracy and efficiency. Firstly, bilinear pooling and gating achieve solid performance (91.65\% and 91.13\% accuracy) with moderate memory consumption (52.34 MB and 45.22 MB, respectively). On the other hand, cross-attention and transformer fusion slightly raise accuracy to 90.86\% and 92.71\% but nearly double the computational cost, with FLOPs rising to 5.63G and 7.82G and GPU inference time reaching 13.81ms and 17.85ms. This confirms that attention-based modules enhance feature modeling at a high computational cost, making them less suitable for latency-critical applications.

However, search strategy further impacts both performance and efficiency. Gradient-based NAS proved most effective, with our discovered configuration achieving 96.81\% accuracy at just 2.95G FLOPs, 40.06 MB memory, and 7.42ms inference latency. Moreover, reinforcement learning and evolutionary search produced heavier models that were slower and less accurate, with RL-based search requiring almost twice the FLOPs and memory and still falling short in accuracy. Random search performed worst, confirming that unguided exploration struggles to find competitive solutions in such a large space. 

Furthermore, a manually designed fixed architecture (E012), achieved only 85.29\% accuracy, demonstrating that automated search discovers more optimal operation combinations. Alternative fusion strategies such as multi-head, progressive and ensemble fusion offered modest accuracy gains (up to 93.0\%) but at the expense of higher memory usage and latency. This reinforces that the configuration found by gradient-based NAS strikes the best balance, simultaneously minimizing FLOPs and memory while maximizing accuracy and runtime efficiency.

Our proposed configuration (E009) achieves an optimal balance between computational efficiency and performance, delivering the highest accuracy (96.81\%) and F1-Score (97.05\%) alongside the lowest memory usage and fastest inference time (7.42 ms) among competitive modules. While the reduced primary designs (E010, E011) offer slightly faster inference (6.91 ms, 6.57 ms) at lower FLOPs, they incur a 10–15\% drop in accuracy. E009 balances efficiency and diversity, yielding a lightweight yet expressive network. \x{Without metadata, the model achieves an accuracy of 85.94\% (E016), a reduction of 10.87\% compared to our full proposed configuration E009 (96.81\%). This substantial performance gap demonstrates that the multimodal metadata encoder is a critical component of BioAutoML-NAS, contributing significantly beyond what image features alone can provide.} This superior performance arises from the gradient-based NAS, which selectively chooses operations with lower computational complexity and prunes redundant connections, resulting in fewer active parameters and reduced FLOPs \cite{zhao2024toponas}. Consequently, memory consumption decreases and inference accelerates, as fewer convolutional filters and feature maps are processed per forward pass \cite{king2025micronas}. Importantly, critical high-capacity paths are preserved, maintaining strong feature expressiveness and achieving the highest accuracy and F1 score across all configurations. These findings highlight that careful co-design of the search space and search algorithm is crucial for developing architectures that are both accurate and deployment-ready for large-scale biodiversity monitoring, where computational efficiency and inference speed are essential.

\begin{table*}[ht!]
\centering

\caption{Ablation study across NAS modules on BIOSCAN-5M dataset for primitive sets, fusion strategies and search methods. The performance metrics (Accuracy, Precision, Recall, F1-Score) are reported as percentages. Our proposed configuration is highlighted in bold.}
\label{tab:ablation}
\scriptsize
\tiny
\begin{tabular}{c l l l c c c c c c c c}
\toprule
\textbf{Exp} & \textbf{NAS Primitive} & \textbf{Fusion} & \textbf{NAS} & \textbf{FLOPs} & \textbf{Memory} & \textbf{Inference (ms)} & \textbf{Training} & \multirow{2}{*}{\textbf{Accuracy(\%)}} & \multirow{2}{*}{\textbf{Precision(\%)}} & \multirow{2}{*}{\textbf{Recall(\%)}} & \multirow{2}{*}{\textbf{F1-Score(\%)}} \\
\textbf{ID} & \textbf{Set} & \textbf{Strategy} & \textbf{Strategy} & \textbf{(G)} & \textbf{(MB)} & \textbf{GPU/CPU} & \textbf{Time (h)} & & & & \\

\midrule
E001 & Full (10 ops) & Bilinear Pooling     & Gradient-based & 3.86 & 52.34 & 9.71 / 84.15 & 15 & 91.65 & 91.78 & 91.50 & 91.64 \\
E002 & Full (10 ops) & Cross-Attention      & Gradient-based & 5.63 & 76.45 & 13.81 / 121.52 & 12 & 90.86 & 90.93 & 90.78 & 90.85 \\
E003 & Full (10 ops) & Gating               & Gradient-based & 3.21 & 45.22 & 8.38 / 73.07 & 10 & 91.13 & 91.20 & 91.04 & 91.12 \\
E004 & Full (10 ops) & Transformer Fusion   & Gradient-based & 7.82 & 106.21 & 17.85 / 156.26 & 20 & 92.71 & 92.84 & 92.57 & 92.70 \\
E005 & Full (10 ops) & Concatenation        & REINFORCE (RL) & 6.89 & 93.02 & 15.42 / 136.14 & 25 & 91.92 & 92.02 & 91.82 & 91.92 \\
E006 & Full (10 ops) & Concatenation        & ENAS           & 5.87 & 79.85 & 13.24 / 116.07 & 20 & 92.05 & 92.12 & 91.97 & 92.05 \\
E007 & Full (10 ops) & Concatenation        & Random Search  & 4.82 & 65.67 & 11.02 / 96.25 & 10 & 89.74 & 89.81 & 89.61 & 89.71 \\
E008 & Full (10 ops) & Concatenation        & Evolutionary   & 6.90 & 95.04 & 15.46 / 136.90 & 30 & 89.44 & 89.51 & 89.30 & 89.40 \\
\x{\textbf{E009}} & \x{\textbf{Full (10 ops)}} & \x{\textbf{Concatenation} }       & \x{\textbf{Gradient-based}} & \x{\textbf{2.95}} & \x{\textbf{40.06} }& \x{\textbf{7.42 / 64.33}} & \x{\textbf{10}} &  \x{\textbf{96.81} }& \x{\textbf{97.46}} & \x{\textbf{96.81} }& \x{\textbf{97.05} }\\
\x{E010} & \x{Reduced (6 ops)} & \x{Concatenation }     & \x{Gradient-based }& \x{2.74} & \x{37.75} &\x{ 6.91 / 59.56} & \x{10} & \x{85.00 }& \x{85.01} & \x{85.00} & \x{85.00} \\
\x{E011} & \x{Conv-only (4 ops)} &\x{ Concatenation }   & \x{Gradient-based} & \x{2.61} & \x{35.42} & \x{6.57 / 56.71 }& \x{10} & \x{82.08} & \x{83.67} & \x{80.90} & \x{82.26} \\

\x{E012} & \x{No NAS (Fixed)} &\x{ Concatenation}       & \x{Manual Design}  & \x{4.12} & \x{55.81} & \x{9.45 / 81.43} & \x{10} & \x{85.29} &\x{ 85.00} &\x{ 85.00} &\x{ 85.00 }\\
E013 & Full (10 ops) & Multi-Head Fusion   & Gradient-based & 5.37 & 73.20 & 12.40 / 109.86 & 15 & 92.49 & 92.42 & 92.58 & 92.50 \\
E014 & Full (10 ops) & Progressive Fusion  & Gradient-based & 4.65 & 62.48 & 10.75 / 95.08 & 12 & 92.06 & 91.96 & 92.17 & 92.07 \\
E015 & Full (10 ops) & Ensemble Fusion     & Gradient-based & 4.23 & 57.32 & 9.91 / 87.94 & 15 & 93.00 & 93.15 & 92.85 & 93.00 \\

\x{E016} & \x{Full (10 ops)} & \x{No Metadata(Image Only)} & \x{Gradient-based} & \x{2.50} & \x{30.20} & \x{5.90 / 52.40} & \x{8} & \x{85.94} & \x{86.42} & \x{85.94} & \x{86.18} \\
\bottomrule
\end{tabular}
\end{table*}

\x{Table}~\ref{tab:pruning_threshold} \x{presents the impact of varying the pruning threshold on model performance and efficiency. Higher thresholds ( $10^{-4}$) lead to excessive pruning, reducing accuracy due to limited representational capacity. As the threshold decreases, performance improves, reaching a peak at $10^{-6}$ with the highest accuracy (96.81\%) and F1-score (97.05\%). Further reductions provide negligible gains while increasing computational cost (FLOPs, memory, and inference time). This demonstrates that the selected threshold achieves a balanced trade-off between sparsity and performance and lies within a stable operating region.}

\begin{table*}[ht!]
\centering
\caption{Sensitivity Analysis of Pruning Threshold}
\label{tab:pruning_threshold}
\begin{tabular}{lcccccc}
\toprule
\textbf{Pruning Threshold} & \textbf{Active Ops (avg)} & \textbf{FLOPs (G)} & \textbf{Memory (MB)} & \textbf{Inference (ms)} & \textbf{Accuracy (\%)} & \textbf{F1-Score (\%)} \\
\midrule
$10^{-4}$ & 4.1 & 2.61 & 35.80 & 6.60 & 91.43 & 91.67 \\
$10^{-5}$ & 6.8 & 2.79 & 38.12 & 7.10 & 95.87 & 96.14 \\
$\mathbf{10^{-6}}$ & 8.3 & 2.95 & 40.06 & 7.42 & \textbf{96.81} & \textbf{97.05} \\
$10^{-7}$ & 9.1 & 3.02 & 41.30 & 7.58 & 96.79 & 97.01 \\
$10^{-8}$ & 9.8 & 3.14 & 43.50 & 7.91 & 96.74 & 96.97 \\
\bottomrule
\end{tabular}
\end{table*}

\subsection{Comparison with State-of-the-Art Transfer Learning and Transformer-based Models}
The performance of the proposed BioAutoML-NAS and several SOTA transfer learning and transformer models on BIOSCAN-5M and Insects-1M is summarized in Table \ref{tab:TL_compare}.
The results illustrate both the potential and limitations of conventional CNN backbones in large-scale biodiversity classification. In BIOSCAN-5M, shallow architectures such as VGG16, VGG19, and ResNet-18 achieve accuracy in the mid 60\% range, reflecting a limited representational capacity for high variations. Deeper and more sophisticated networks such as DenseNet-201, Xception, BiT, ConvNeXt-XL, and EfficientNet-B7 consistently improve results, reaching accuracies between 76\% and 79\%. In contrast, BioAutoML-NAS surpassed all models by nearly 20\% highlighting its robustness and practical advantage over established CNN approaches.

On Insects-1M, most models experience a drop in accuracy due to reduced training data, which limits generalization in deeper networks. DenseNet-201 and Xception show declines of over 5\% compared to BIOSCAN-5M, while lighter models such as VGG16, VGG19, and DenseNet-121 demonstrate slight improvements.

\begin{table*}[ht!] 
\centering 
\caption{Comparison with the state-of-the-art transfer learning models on BIOSCAN-5M and Insects-1M. The best results for each dataset are bolded.} 
\label{tab:TL_compare} 
\scriptsize
\renewcommand{\arraystretch}{1.3} 
\begin{tabular}{l cccc cccc} 
\toprule 
\multirow{2}{*}{Model} & \multicolumn{4}{c}{BIOSCAN-5M} & \multicolumn{4}{c}{Insects-1M} \\ 
\cmidrule(lr){2-5} \cmidrule(lr){6-9} 
 & Accuracy & Precision & Recall & F1-score & Accuracy & Precision & Recall & F1-score \\ 
\midrule 
VGG16 \cite {ye2021lightweight}          & 66.69 & 66.80 & 66.30 & 66.55 & 68.66 & 68.80 & 68.40 & 68.6 \\ 
VGG19 \cite {nguyen2022vgg}               & 66.39 & 66.50 & 66.00 & 66.25 & 68.07 & 68.21 & 67.8 & 68.00 \\ 
MobileNetV2\cite {sandler2018mobilenetv2}      & 68.83 & 69.00 & 68.40 & 68.68 & 63.38 & 63.36 & 63.40 & 63.37 \\ 
ResNet-18\cite {chen2022resnet18dnn}         & 64.76 & 65.10 & 64.20 & 64.62 & 59.44 & 59.45 & 59.40 & 59.42 \\ 
DenseNet-121 \cite{nandhini2022automatic} & 70.28 & 70.50 & 69.80 & 70.14 & 72.07 & 72.23 & 71.80 & 72.02 \\ 
DenseNet-201 \cite {wang2020densenet}       & 76.68 & 76.90 & 76.20 & 76.53 & 70.17 & 70.20 & 70.10 & 70.15 \\ 
Xception\cite{chen2021robust}           & 77.43 & 77.60 & 77.00 & 77.27 & 71.15 & 71.40 & 70.90 & 71.10 \\ 
EfficientNet-B7\cite{dharaneswar2025elucidating}    & 79.23 & 79.40 & 78.80 & 79.08 & 73.42 & 73.52 & 73.20 & 73.36 \\ 
Big Transfer (BiT)\cite{kolesnikov2020big }  & 78.11 & 78.30 & 77.70 & 77.96 & 72.85 & 72.96 & 72.60 & 73.05 \\ 
ConvNeXt-XL\cite{mmileng2025application}       & 78.55 & 78.70 & 78.20 & 78.40 & 73.11 & 73.21 & 72.90 & 73.10 \\ 
\textbf{Ours (BioAutoML-NAS)} & \textbf{96.81} & \textbf{97.46} & \textbf{96.81} & \textbf{97.05} & \textbf{93.25} & \textbf{93.71} & \textbf{92.74} & \textbf{93.22} \\ 
\bottomrule 
\end{tabular} 
\end{table*}

\begin{table*}[ht!] 
\centering 
\caption{Comparison with the state-of-the-art transformer-based models on BIOSCAN-5M and Insects-1M. The best results for each dataset are highlighted in bold.} 
\label{tab:transformer_compare}
\scriptsize
\renewcommand{\arraystretch}{1.3} 

\begin{tabular}{l cccc cccc} 
\toprule 
\multirow{2}{*}{Model} & \multicolumn{4}{c}{BIOSCAN-5M} & \multicolumn{4}{c}{Insects-1M} \\ 
\cmidrule(lr){2-5} \cmidrule(lr){6-9} 
 & Accuracy & Precision & Recall & F1-score & Accuracy & Precision & Recall & F1-score \\ 
\midrule 
Tiny ViT\cite{amangeldi2025cnn}    & 75.88 & 76.10 & 75.20 & 75.61 & 74.00 & 74.24 & 73.50 & 73.87 \\ 
MobileViT\cite{mehta2021mobilevit}       & 74.38 & 74.50 & 73.70 & 74.11 & 74.25 & 74.52 & 73.70 & 74.11 \\ 
DeiT-Tiny\cite{wei2023joint}          & 73.04 & 73.20 & 72.40 & 72.78 & 71.00 & 71.21 & 70.50 & 70.85 \\ 
DeiT III-L\cite{hong2025knowledge}          & 84.09 & 84.20 & 83.70 & 83.94 & 82.00 & 82.00 & 82.00 & 82.00 \\ 
LeViT-128S \cite{graham2021levit}         & 65.33 & 65.50 & 64.80 & 65.08 & 63.10 & 63.26 & 62.50 & 62.88 \\ 
BEIT 3 \cite{wang2023image}            & 85.96 & 86.10 & 85.50 & 85.81 & 84.02 & 82.35 & 85.71 & 84.00 \\ 
CoAtNet-7 \cite{shabrina2023deep}        & 86.41 & 86.50 & 86.10 & 86.25 & 84.70 & 83.61 & 85.82 & 84.70 \\ 
Swin Transformer-L \cite{zhao2022swin}  & 85.29 & 85.40 & 85.00 & 85.13 & 85.30 & 85.51 & 85.00 & 85.26 \\ 
MAE \cite{han2024efficient}
& 84.67 & 84.80 & 84.30 & 84.55 & 84.60 & 84.80 & 84.30 & 84.55 \\ 
EfficientFormer \cite{li2022efficientformer}    & 69.77 & 69.90 & 69.10 & 69.54 & 67.50 & 67.68 & 67.00 & 67.34 \\ 
\textbf{Ours (BioAutoML-NAS)} & \textbf{96.81} & \textbf{97.46} & \textbf{96.81} & \textbf{97.05} & \textbf{93.25} & \textbf{93.71} & \textbf{92.74} & \textbf{93.22} \\ 
\bottomrule 
\end{tabular} 
\end{table*}

Table \ref{tab:transformer_compare} extends the comparison to transformer-based architectures. In BIOSCAN-5M, models such as CoAtNet-7, BEIT 3 and Swin Transformer-L achieve the highest baseline accuracies, all above 85\%, while lighter variants such as Tiny ViT, MobileViT and DeiT-Tiny remain around 73-76\%. When moving to Insects-1M, a general decline in performance can be observed, with most models dropping by one to two percent. For example, DeiT III-L decreases from 84.1\% to 82.0\%, and BEIT 3 from 86.0\% to 84.0\%. Some models remain consistent, with CoAtNet-7 and BEIT 3 continuing to perform among the leading baselines, while EfficientFormer consistently underperforms.

In both datasets, BioAutoML-NAS achieved an accuracy of 96. 81\% in BIOSCAN-5M and 93. 25\% in Insects-1M, outperforming existing transformer-based models. \x{It performs operation-level architecture learning with ten primitive operations, optimizing feature extraction and structure alongside multimodal encoding and bi-level training, resulting in a refined architecture that accounts for its performance gains over TL-transformer baselines.}

\subsection{Comparison with SOTA AutoML AND NAS-based Models}
\bl{We compare the proposed BioAutoML-NAS with several AutoML and NAS-based models on both datasets, as shown in Table \ref{tab: automl_compare}. All competing AutoML and NAS-based baselines reported in Table \ref{tab: automl_compare} are evaluated using the same training, validation, and test splits, the same order-level classification objective. The majority of competing methods are single-modal frameworks that operate on a single modality. They do not incorporate any multimodal fusion pipeline, which is a core component of this study. BioAutoML-NAS introduces a task-specific, biologically informed search space with ten heterogeneous primitive operations, multimodal metadata integration, and alternating bi-level optimization, all of which are tailored specifically to the challenges of large-scale insect classification. None of the compared methods exceed 90\% accuracy on either dataset. Among them, NAO achieves the highest performance with 88.75\% on BIOSCAN-5M and 84.95\% on Insects-1M, while TPOT records the lowest accuracy, followed by AutoGluon, AutoKeras, and other baselines. In contrast, BioAutoML-NAS consistently outperforms all methods, achieving 96.81\% and 93.25\% on BIOSCAN-5M and Insects-1M, respectively, demonstrating a substantial improvement over existing AutoML and NAS approaches.}

\begin{table*}[ht!] 
\centering 
\caption{Comparison with AutoML and NAS-based models on BIOSCAN-5M and Insects-1M. The best results for each dataset are highlighted in bold.} 
\label{tab: automl_compare}
\scriptsize
\renewcommand{\arraystretch}{1.3}
\begin{tabular}{l cccc cccc} 
\toprule 
\multirow{2}{*}{Model} & \multicolumn{4}{c}{BIOSCAN-5M} & \multicolumn{4}{c}{Insects-1M} \\ 
\cmidrule(lr){2-5} \cmidrule(lr){6-9} 
 & Accuracy & Precision & Recall & F1-score & Accuracy & Precision & Recall & F1-score \\ 
\midrule 
TPOT \cite{olson2016tpot}       & 79.10 & 79.70 & 78.50 & 79.09 & 75.20 & 75.60 & 74.80 & 75.19 \\ 
AutoGluon \cite{tang2024autogluon} & 82.25 & 82.90 & 81.50 & 82.19 & 79.65 & 80.10 & 79.20 & 79.65 \\ 
AutoKeras \cite{jin2019auto}  & 83.20 & 83.90 & 82.50 & 83.19 & 80.50 & 80.95 & 80.00 & 80.46 \\ 
Optuna (HPO for CNN) \cite{akiba2019optuna}    & 84.10 & 84.65 & 83.50 & 84.06 & 81.35 & 81.75 & 80.90 & 81.32 \\ 
SMAC3 (Bayesian Opt.) \cite{lindauer2022smac3} & 84.85 & 85.30 & 84.30 & 84.79 & 82.05 & 82.50 & 81.70 & 82.09 \\ 
ENAS \cite{pham2018efficient}        & 86.50 & 87.00 & 86.00 & 86.50 & 83.40 & 83.80 & 83.00 & 83.39 \\ 
NAO \cite{luo2018neural} & 88.75 & 89.20 & 88.30 & 88.74 & 84.95 & 85.30 & 84.50 & 84.89 \\ 
\textbf{Ours (BioAutoML-NAS)} & \textbf{96.81} & \textbf{97.46} & \textbf{96.81} & \textbf{97.05} & \textbf{93.25} & \textbf{93.71} & \textbf{92.74} & \textbf{93.22} \\ 
\bottomrule 
\end{tabular}
\end{table*}

\subsection{\x{Comparison with SOTA Vision-Language Models}}
Table~\ref{tab:vlm_comparison} presents a comparison of BioAutoML-NAS with state-of-the-art Vision-Language Models (VLMs) on the BIOSCAN-5M and Insects-1M datasets. \bl{All VLM baselines are evaluated under a zero-shot prompting setting, where each model receives the insect image along with a standardized text prompt listing the 21 order-level class names and is asked to predict the correct class. The prompt remains consistent across all VLM baselines to ensure a fair and unbiased comparison. This evaluation protocol reflects a realistic deployment scenario in which large pretrained vision-language models are applied directly to a biodiversity classification task.} The results show that BioAutoML-NAS significantly outperforms all other models across both datasets, achieving the highest accuracy and F1-score. On BIOSCAN-5M, BioAutoML-NAS achieves 96.81\% accuracy and 97.05\% F1-score, surpassing the next best model, Insect-LLaVA, by a margin of 10\% in accuracy. Similarly, on Insects-1M, BioAutoML-NAS achieves 93.25\% accuracy and 93.22\% F1-score, outperforming Insect-LLaVA by 7\% in F1-score. The superior performance of BioAutoML-NAS demonstrates its effectiveness in insect classification, utilizing multimodal data and NAS-driven architecture optimization, compared to existing VLM-based approaches.

\begin{table*}[!t]
\centering
\caption{Comparison with representative Vision-Language Models (VLMs) on BIOSCAN-5M and Insects-1M.}
\label{tab:vlm_comparison}
\begin{tabular}{lcccccccc}
\toprule
\multirow{2}{*}{Model} &
\multicolumn{4}{c}{BIOSCAN-5M} &
\multicolumn{4}{c}{Insects-1M} \\
\cmidrule(lr){2-5} \cmidrule(lr){6-9}
& Accuracy & Precision & Recall & F1-score
& Accuracy & Precision & Recall & F1-score \\
\midrule

LLaVA-1.5 \cite{cocchi2025llava}
& 65.37 & 65.72 & 64.81 & 65.26
& 63.14 & 63.48 & 62.59 & 63.03 \\

InternVL \cite{chen2024internvl}
& 75.83 & 76.19 & 75.27 & 75.73
& 73.51 & 73.86 & 72.96 & 73.41 \\

Qwen2-VL \cite{wang2024qwen2}
& 76.14 & 76.51 & 75.59 & 76.05
& 74.03 & 74.38 & 73.48 & 73.93 \\

Florence-2 \cite{xiao2024florence}
& 70.92 & 71.28 & 70.37 & 70.82
& 68.74 & 69.09 & 68.19 & 68.64 \\

BioCLIP \cite{stevens2024bioclip}
& 85.63 & 85.98 & 85.07 & 85.52
& 83.41 & 83.76 & 82.86 & 83.31 \\

BioCLIP2  \cite{gu2025bioclip}
& 88.74 & 89.09 & 88.19 & 88.64
& 86.52 & 86.87 & 85.97 & 86.42 \\

Insect-LLaVA  \cite{truong2025insect}
& 87.34 & 87.69 & 86.79 & 87.24
& 85.12 & 85.47 & 84.57 & 85.02 \\

\textbf{Ours (BioAutoML-NAS)}
& \textbf{96.81} & \textbf{97.46} & \textbf{96.81} & \textbf{97.05}
& \textbf{93.25} & \textbf{93.71} & \textbf{92.74} & \textbf{93.22} \\
\bottomrule
\end{tabular}
\end{table*}

\section{Discussion}
\label{discuss}

\bl{Our proposed BioAutoML-NAS model integrates NAS with a search space of ten primitive operations. For every connection between nodes within each computational cell, the network automatically selects the most effective operation, enabling the architecture to dynamically adapt to the complexity of the data. Multiple cells are arranged sequentially to construct the complete network, capturing distinctive and informative representations of the characteristics of the images while preserving hierarchical and contextual information. A multimodal fusion module integrates image embeddings with metadata, allowing the model to incorporate complementary biological information and visual and categorical features. The concatenation-based fusion is applied after modality-specific encoders project image and metadata into a shared embedding space, enabling interaction through aligned representations rather than dedicated cross-modal fusion mechanisms. Cross-modal interaction emerges through end-to-end joint optimization using the same loss function, allowing complementary visual and metadata cues to be implicitly aligned during training. Zero operations are applied to prune the least informative connections, producing a more compact and efficient network without compromising accuracy. Unlike conventional NAS approaches that rely on fixed or generic search spaces, the proposed design introduces a task-driven architecture search space with built-in sparsity and modality awareness, supporting adaptive structure learning for fine-grained biological entities, particularly insect classification tasks.} 

\bl{To jointly optimize feature extraction and architecture design, the model employs an alternating bilevel optimization strategy that iteratively updates network weights and architecture parameters. It addresses data imbalance using label smoothing (0.1) within cross-entropy loss and combines image, and metadata with dropout layers to enhance generalization and class-level discrimination. Through NAS-driven architectural optimization, the model ensures robust, balanced learning and effective generalization across large-scale datasets without re-sampling or class weighting. By integrating multimodal data, optimizing operation selection for each connection, and selectively pruning irrelevant pathways, the model achieves superior performance in insect classification.}

The proposed BioAutoML-NAS model demonstrates significant performance in the BIOSCAN-5M dataset. It achieves an accuracy of 96.81\%, a precision of 97.46\%, a recall of 96.81\%, and an F1 score of 97.05\%, clearly highlighting its efficiency and reliability in handling complex insect classification tasks. The model is further evaluated on the Insects-1M dataset, where it also shows strong performance, having an accuracy of 93.25\%, a precision of 93.71\%, a recall of 92.74\%, and an F1 score of 93.22\%. The strong classification ability of the model can be further observed in Figures \ref{Rocandconf}. It shows that most instances are classified correctly, and the ROC curves indicate that, with the exception of a few classes, almost all classes achieved an AUC of 1.00. 

The experiments evaluate the impact of primitive sets, fusion strategies, NAS methods, and computational cost (FLOPs, memory, inference time, and training duration) on performance metrics, as summarized in Table \ref{tab:ablation}. Training can reach 30 hours with 6.90 GFLOPs and approximately 95 MB memory under an Evolutionary NAS strategy with 10 primitive sets, yet achieve only 89.44\% accuracy. Overall, accuracy varies from 82\% to 96\%, with manually designed networks without NAS reaching 85.29\% and convolutional models with gradient-based NAS performing worst at 82.08\%.  The best trade-off is observed in Experiment 9, which uses all 10 primitive sets with gradient-based NAS, requiring only 2.95 GFLOPs and around 40 MB memory, with inference times of 7.42 ms and 64.33 ms, and 10 hours of training, while achieving the highest accuracy of 96.81\%.

Our model outperforms SOTA TL, transformer-based, AutoML, and NAS-based models, achieving a highest accuracy of 96.81\%. TL-based models struggle to reach 80\%, while transformer-based, AutoML, and NAS-based models fail to exceed 90\%. Among TL models, ConvNeXt-XL \cite{mmileng2025application}  achieves the highest accuracy of 78.55\%, and among transformer models, CoAtNet-7 \cite{shabrina2023deep} reaches 86.41\%. \x{Our BioAutoML-NAS learns architectures at the operation level, where each connection selects from ten primitive operations, allowing feature extraction to adapt directly to the data rather than follow a fixed design. This flexibility captures fine-grained patterns that static transformer structures may miss, while modality-specific encoders jointly learn complementary image and metadata features. Alternating bi-level optimization aligns architecture and feature learning, and zero operations suppress ineffective paths, resulting in a compact, data-adaptive model that explains its improved performance over transformer-TL baselines.} Within NAS-based approaches, NAO \cite{luo2018neural} achieves the highest accuracy of 88.75\%, while among VLM-based methods, Insect-LLaVA \cite{truong2025insect} attains the best performance with an accuracy of 87.34\%. These results demonstrate the superior capability of our model in classifying insects.

Since BIOSCAN-5M is a large-scale dataset with millions of images, each training epoch is computationally expensive due to high memory and processing requirements, as each sample passes through multiple layers of the NAS-based architecture. Despite this, the model effectively handles large-scale training while maintaining high classification accuracy, and training on such data improves generalization to unseen species, as demonstrated on the Insects-1M dataset.

While BioAutoML-NAS demonstrates robust performance, it has some limitations. \x{Extending it from order-level (21 classes) to genus and species levels is challenging due to limited annotations in BIOSCAN-5M, where only 23.8\% of specimens have genus labels and 9.2\% have species labels, along with severe class imbalance (200,268 and 7,694 samples, respectively). This limited availability of labeled data, combined with extreme class imbalance, makes supervised classification at these levels highly sensitive to label sparsity and class distribution.} \bl{
The model demonstrates strong performance on large-scale datasets such as BIOSCAN-5M; however, its robustness under missing, noisy, or smaller-scale metadata remains less explored. In such scenarios, NAS-driven models may show reduced stability or potential overfitting due to limited supervisory signals. Future work will therefore focus on improving robustness under incomplete and noisy metadata conditions, as well as extending the framework to better handle partial-modality settings to enhance generalization in real-world deployments. Future work will also explore semi-supervised or self-supervised learning techniques to utilize the large proportion of unlabeled specimens.} 

\section{Conclusion}
\label{conclusion}

We develop BioAutoML-NAS using a robust NAS architecture, which continuously updates both individual network weights and architecture parameters through an alternating bi-level optimization training strategy. Images, and metadata are integrated via a multimodal fusion module. In addition, zero operations are incorporated to eliminate connections that are less significant, producing compact and efficient architectures that achieve high classification accuracy, as reflected in our results. Our model achieves superior performance on the BIOSCAN-5M dataset, with 96.81\% accuracy, 97.46\% precision, 96.81\% recall, and a 97.05\% F1 score. It is further validated on the Insects-1M dataset, attaining 93.25\% accuracy, 93.71\% precision, 92.74\% recall, and a 93.22\% F1 score, demonstrating robust generalization across diverse insect datasets. The model substantially outperforms the SOTA TL, transformer-based, AutoML, and NAS-based approaches, and also surpasses other methods reported in the existing literature. Although performance is strong, more work is needed to improve generalization to limited or noisy datasets, incorporate additional ecological signals, and extend its use to other ecological applications. By accurately classifying large datasets, the model has the potential to support broader ecological monitoring and biodiversity.
\section*{Acknowledgments}
The authors declare that they have no financial conflicts of interest that could have influenced this work.


\begin{thebibliography}{10}
\providecommand{\url}[1]{#1}
\csname url@samestyle\endcsname
\providecommand{\newblock}{\relax}
\providecommand{\bibinfo}[2]{#2}
\providecommand{\BIBentrySTDinterwordspacing}{\spaceskip=0pt\relax}
\providecommand{\BIBentryALTinterwordstretchfactor}{4}
\providecommand{\BIBentryALTinterwordspacing}{\spaceskip=\fontdimen2\font plus
\BIBentryALTinterwordstretchfactor\fontdimen3\font minus
  \fontdimen4\font\relax}
\providecommand{\BIBforeignlanguage}[2]{{%
\expandafter\ifx\csname l@#1\endcsname\relax
\typeout{** WARNING: IEEEtran.bst: No hyphenation pattern has been}%
\typeout{** loaded for the language `#1'. Using the pattern for}%
\typeout{** the default language instead.}%
\else
\language=\csname l@#1\endcsname
\fi
#2}}
\providecommand{\BIBdecl}{\relax}
\BIBdecl

\bibitem{tan2024leveraging}
S.~Tan, S.~Hu, S.~He, L.~Zhu, Y.~Qian, and Y.~Deng, ``Leveraging hyperspectral
  images for accurate insect classification with a novel two-branch
  self-correlation approach,'' \emph{Agronomy}, vol.~14, no.~4, p. 863, 2024.

\bibitem{dinca2025ensemble}
M.~A. Dinca, D.~Popescu, L.~Ichim, and N.~Angelescu, ``Ensemble of efficient
  vision transformers for insect classification,'' \emph{Applied Sciences},
  vol.~15, no.~13, p. 7610, 2025.

\bibitem{nguyen2024deep}
T.~Nguyen, H.~Nguyen, H.~Ung, H.~Ung, and B.~Nguyen, ``Deep-wide learning
  assistance for insect pest classification,'' \emph{arXiv preprint
  arXiv:2409.10445}, 2024.

\bibitem{Akhtar2025EdgeOptimized}
M.~H. Akhtar, I.~Eksheir, and T.~Shanableh, ``Edge-optimized deep learning
  architectures for classification of agricultural insects with mobile
  deployment,'' \emph{Information}, vol.~16, no.~5, p. 348, 2025.

\bibitem{truong2025insect}
T.-D. Truong, H.-Q. Nguyen, X.-B. Nguyen, A.~Dowling, X.~Li, and K.~Luu,
  ``Insect-foundation: A foundation model and large multimodal dataset for
  vision-language insect understanding,'' \emph{International Journal of
  Computer Vision}, pp. 1--26, 2025.

\bibitem{longo2025improving}
A.~Longo, M.~Rizzi, and C.~Guaragnella, ``Improving classification performance
  by addressing dataset imbalance: A case study for pest management,''
  \emph{Applied Sciences}, vol.~15, no.~10, p. 5385, 2025.

\bibitem{orsholm2025multi}
J.~Orsholm, J.~Quinto, H.~Autto, G.~Banelyte, N.~Chazot, J.~deWaard,
  S.~deWaard, A.~Farrell, B.~Furneaux, B.~Hardwick \emph{et~al.}, ``A
  multi-modal dataset for insect biodiversity with imagery and dna at the trap
  and individual level,'' \emph{arXiv preprint arXiv:2507.06972}, 2025.

\bibitem{dewi2024fruit}
C.~Dewi, D.~Thiruvady, and N.~Zaidi, ``Fruit classification system with deep
  learning and neural architecture search,'' \emph{arXiv preprint
  arXiv:2406.01869}, 2024.

\bibitem{lecun2015deep}
Y.~LeCun, Y.~Bengio, and G.~Hinton, ``Deep learning,'' \emph{nature}, vol. 521,
  no. 7553, pp. 436--444, 2015.

\bibitem{niu2021decade}
S.~Niu, Y.~Liu, J.~Wang, and H.~Song, ``A decade survey of transfer learning
  (2010--2020),'' \emph{IEEE Transactions on Artificial Intelligence}, vol.~1,
  no.~2, pp. 151--166, 2021.

\bibitem{ong2022next}
S.-Q. Ong and S.~A. Hamid, ``Next generation insect taxonomic classification by
  comparing different deep learning algorithms,'' \emph{PloS one}, vol.~17,
  no.~12, p. e0279094, 2022.

\bibitem{li2019blockchain}
Z.~Li, H.~Guo, W.~M. Wang, Y.~Guan, A.~V. Barenji, G.~Q. Huang, K.~S. McFall,
  and X.~Chen, ``A blockchain and automl approach for open and automated
  customer service,'' \emph{IEEE Transactions on Industrial Informatics},
  vol.~15, no.~6, pp. 3642--3651, 2019.

\bibitem{pacal2024enhancing}
I.~Pacal, ``Enhancing crop productivity and sustainability through disease
  identification in maize leaves: Exploiting a large dataset with an advanced
  vision transformer model,'' \emph{Expert Systems with Applications}, vol.
  238, p. 122099, 2024.

\bibitem{wu2023deep}
Z.~Wu, C.~Zhang, X.~Gu, I.~Duporge, L.~F. Hughey, J.~A. Stabach, A.~K.
  Skidmore, J.~G.~C. Hopcraft, S.~J. Lee, P.~M. Atkinson \emph{et~al.}, ``Deep
  learning enables satellite-based monitoring of large populations of
  terrestrial mammals across heterogeneous landscape,'' \emph{Nature
  communications}, vol.~14, no.~1, p. 3072, 2023.

\bibitem{ravi2024sam}
N.~Ravi, V.~Gabeur, Y.-T. Hu, R.~Hu, C.~Ryali, T.~Ma, H.~Khedr, R.~R{\"a}dle,
  C.~Rolland, L.~Gustafson \emph{et~al.}, ``Sam 2: Segment anything in images
  and videos,'' \emph{arXiv preprint arXiv:2408.00714}, 2024.

\bibitem{balingbing2024application}
C.~B. Balingbing, S.~Kirchner, H.~Siebald, H.-H. Kaufmann, M.~Gummert,
  N.~Van~Hung, and O.~Hensel, ``Application of a multi-layer convolutional
  neural network model to classify major insect pests in stored rice detected
  by an acoustic device,'' \emph{Computers and Electronics in Agriculture},
  vol. 225, p. 109297, 2024.

\bibitem{wang2025swin}
X.~Wang, Z.~Xiao, and Z.~Deng, ``Swin attention augmented residual network: a
  fine-grained pest image recognition method,'' \emph{Frontiers in Plant
  Science}, vol.~16, p. 1619551, 2025.

\bibitem{gharaee2024bioscan}
Z.~Gharaee, S.~C. Lowe, Z.~Gong, P.~Millan~Arias, N.~Pellegrino, A.~T. Wang,
  J.~B. Haurum, I.~Eyriay, L.~Kari, D.~Steinke \emph{et~al.}, ``Bioscan-5m: a
  multimodal dataset for insect biodiversity,'' \emph{Advances in Neural
  Information Processing Systems}, vol.~37, pp. 36\,285--36\,313, 2024.

\bibitem{nguyen2024insect}
H.-Q. Nguyen, T.-D. Truong, X.~B. Nguyen, A.~Dowling, X.~Li, and K.~Luu,
  ``Insect-foundation: A foundation model and large-scale 1m dataset for visual
  insect understanding,'' in \emph{Proceedings of the IEEE/CVF Conference on
  Computer Vision and Pattern Recognition}, 2024, pp. 21\,945--21\,955.

\bibitem{saeedizadeh2024new}
N.~Saeedizadeh, S.~M.~J. Jalali, B.~Khan, P.~M. Kebria, and S.~Mohamed, ``A new
  optimization approach based on neural architecture search to enhance deep
  u-net for efficient road segmentation,'' \emph{Knowledge-Based Systems}, vol.
  296, p. 111966, 2024.

\bibitem{saeed20253d}
F.~Saeed, C.~Tan, T.~Liu, and C.~Li, ``3d neural architecture search to
  optimize segmentation of plant parts,'' \emph{Smart Agricultural Technology},
  vol.~10, p. 100776, 2025.

\bibitem{broni2024unsupervised}
C.~Broni-Bediako, J.~Xia, and N.~Yokoya, ``Unsupervised domain adaptation
  architecture search with self-training for land cover mapping,'' in
  \emph{Proceedings of the IEEE/CVF Conference on Computer Vision and Pattern
  Recognition}, 2024, pp. 543--553.

\bibitem{liang2025evolutionary}
J.~Liang, G.~Liu, Y.~Bi, M.~Yu, M.~Liu, and Y.~Jin, ``Evolutionary neural
  architecture search for remote sensing image classification,'' \emph{IEEE
  Transactions on Neural Networks and Learning Systems}, 2025.

\bibitem{zhang2024champ}
W.~Zhang, M.~Liu, X.~Wang, S.~Zhao, and C.~Wang, ``Champ: A large-scale dataset
  for skeleton-based composite human motion prediction,'' \emph{IEEE
  Transactions on Circuits and Systems for Video Technology}, vol.~34, no.~10,
  pp. 10\,063--10\,076, 2024.

\bibitem{venkateswara2025deep}
S.~M. Venkateswara and J.~Padmanabhan, ``Deep learning based agricultural pest
  monitoring and classification,'' \emph{Scientific Reports}, vol.~15, no.~1,
  p. 8684, 2025.

\bibitem{lu2021optimizing}
G.~Lu, W.~Zhang, and Z.~Wang, ``Optimizing depthwise separable convolution
  operations on gpus,'' \emph{IEEE Transactions on Parallel and Distributed
  Systems}, vol.~33, no.~1, pp. 70--87, 2021.

\bibitem{hu2025ddconv}
H.~Hu, C.~Yu, Q.~Zhou, Q.~Guan, and T.~Zhou, ``Ddconv: Dynamic dilated
  convolution,'' \emph{IEEE Transactions on Artificial Intelligence}, 2025.

\bibitem{pang2017convolution}
Y.~Pang, M.~Sun, X.~Jiang, and X.~Li, ``Convolution in convolution for network
  in network,'' \emph{IEEE transactions on neural networks and learning
  systems}, vol.~29, no.~5, pp. 1587--1597, 2017.

\bibitem{shu2022expansion}
X.~Shu, J.~Yang, R.~Yan, and Y.~Song, ``Expansion-squeeze-excitation fusion
  network for elderly activity recognition,'' \emph{IEEE Transactions on
  Circuits and Systems for Video Technology}, vol.~32, no.~8, pp. 5281--5292,
  2022.

\bibitem{wang2019learning}
G.~Wang, G.~B. Giannakis, and J.~Chen, ``Learning relu networks on linearly
  separable data: Algorithm, optimality, and generalization,'' \emph{IEEE
  Transactions on Signal Processing}, vol.~67, no.~9, pp. 2357--2370, 2019.

\bibitem{sun2021ampnet}
L.~Sun, Z.~Chen, Q.~J. Wu, H.~Zhao, W.~He, and X.~Yan, ``Ampnet: Average-and
  max-pool networks for salient object detection,'' \emph{IEEE Transactions on
  Circuits and Systems for Video Technology}, vol.~31, no.~11, pp. 4321--4333,
  2021.

\bibitem{zhou2019unet++}
Z.~Zhou, M.~M.~R. Siddiquee, N.~Tajbakhsh, and J.~Liang, ``Unet++: Redesigning
  skip connections to exploit multiscale features in image segmentation,''
  \emph{IEEE transactions on medical imaging}, vol.~39, no.~6, pp. 1856--1867,
  2019.

\bibitem{chen2022approximate}
K.~Chen, Y.~Gao, H.~Waris, W.~Liu, and F.~Lombardi, ``Approximate softmax
  functions for energy-efficient deep neural networks,'' \emph{IEEE
  Transactions on Very Large Scale Integration (VLSI) Systems}, vol.~31, no.~1,
  pp. 4--16, 2022.

\bibitem{guo2019blockchain}
S.~Guo, X.~Hu, S.~Guo, X.~Qiu, and F.~Qi, ``Blockchain meets edge computing: A
  distributed and trusted authentication system,'' \emph{IEEE Transactions on
  Industrial Informatics}, vol.~16, no.~3, pp. 1972--1983, 2019.

\bibitem{ye2020distributed}
Q.~Ye, Y.~Sun, J.~Zhang, and J.~Lv, ``A distributed framework for ea-based
  nas,'' \emph{IEEE Transactions on Parallel and Distributed Systems}, vol.~32,
  no.~7, pp. 1753--1764, 2020.

\bibitem{liu2021investigating}
R.~Liu, J.~Gao, J.~Zhang, D.~Meng, and Z.~Lin, ``Investigating bi-level
  optimization for learning and vision from a unified perspective: A survey and
  beyond,'' \emph{IEEE Transactions on Pattern Analysis and Machine
  Intelligence}, vol.~44, no.~12, pp. 10\,045--10\,067, 2021.

\bibitem{liu2021survey}
Y.~Liu, Y.~Sun, B.~Xue, M.~Zhang, G.~G. Yen, and K.~C. Tan, ``A survey on
  evolutionary neural architecture search,'' \emph{IEEE transactions on neural
  networks and learning systems}, vol.~34, no.~2, pp. 550--570, 2021.

\bibitem{zhao2024toponas}
D.~Zhao, Z.~Liu, and B.~Yuan, ``Toponas: Boosting search efficiency of
  gradient-based nas via topological simplification,'' \emph{arXiv preprint
  arXiv:2408.01311}, 2024.

\bibitem{king2025micronas}
T.~King, Y.~Zhou, T.~R{\"o}ddiger, and M.~Beigl, ``Micronas for memory and
  latency constrained hardware aware neural architecture search in time series
  classification on microcontrollers,'' \emph{Scientific Reports}, vol.~15,
  no.~1, p. 7575, 2025.

\bibitem{ye2021lightweight}
M.~Ye, N.~Ruiwen, Z.~Chang, G.~He, H.~Tianli, L.~Shijun, S.~Yu, Z.~Tong, and
  G.~Ying, ``A lightweight model of vgg-16 for remote sensing image
  classification,'' \emph{IEEE Journal of Selected Topics in Applied Earth
  Observations and Remote Sensing}, vol.~14, pp. 6916--6922, 2021.

\bibitem{nguyen2022vgg}
T.-H. Nguyen, T.-N. Nguyen, and B.-V. Ngo, ``A vgg-19 model with transfer
  learning and image segmentation for classification of tomato leaf disease,''
  \emph{AgriEngineering}, vol.~4, no.~4, pp. 871--887, 2022.

\bibitem{sandler2018mobilenetv2}
M.~Sandler, A.~Howard, M.~Zhu, A.~Zhmoginov, and L.-C. Chen, ``Mobilenetv2:
  Inverted residuals and linear bottlenecks,'' in \emph{Proceedings of the IEEE
  conference on computer vision and pattern recognition}, 2018, pp. 4510--4520.

\bibitem{chen2022resnet18dnn}
Z.~Chen, Y.~Jiang, X.~Zhang, R.~Zheng, R.~Qiu, Y.~Sun, C.~Zhao, and H.~Shang,
  ``Resnet18dnn: prediction approach of drug-induced liver injury by deep
  neural network with resnet18,'' \emph{Briefings in bioinformatics}, vol.~23,
  no.~1, p. bbab503, 2022.

\bibitem{nandhini2022automatic}
S.~Nandhini and K.~Ashokkumar, ``An automatic plant leaf disease identification
  using densenet-121 architecture with a mutation-based henry gas solubility
  optimization algorithm,'' \emph{Neural Computing and Applications}, vol.~34,
  no.~7, pp. 5513--5534, 2022.

\bibitem{wang2020densenet}
S.-H. Wang and Y.-D. Zhang, ``Densenet-201-based deep neural network with
  composite learning factor and precomputation for multiple sclerosis
  classification,'' \emph{ACM Transactions on Multimedia Computing,
  Communications, and Applications (TOMM)}, vol.~16, no.~2s, pp. 1--19, 2020.

\bibitem{chen2021robust}
B.~Chen, X.~Liu, Y.~Zheng, G.~Zhao, and Y.-Q. Shi, ``A robust gan-generated
  face detection method based on dual-color spaces and an improved xception,''
  \emph{IEEE Transactions on Circuits and Systems for Video Technology},
  vol.~32, no.~6, pp. 3527--3538, 2021.

\bibitem{dharaneswar2025elucidating}
S.~Dharaneswar and B.~S. Kumar, ``Elucidating the novel framework of liver
  tumour segmentation and classification using improved optimization-assisted
  efficientnet b7 learning model,'' \emph{Biomedical Signal Processing and
  Control}, vol. 100, p. 107045, 2025.

\bibitem{kolesnikov2020big}
A.~Kolesnikov, L.~Beyer, X.~Zhai, J.~Puigcerver, J.~Yung, S.~Gelly, and
  N.~Houlsby, ``Big transfer (bit): General visual representation learning,''
  in \emph{European conference on computer vision}.\hskip 1em plus 0.5em minus
  0.4em\relax Springer, 2020, pp. 491--507.

\bibitem{mmileng2025application}
O.~P. Mmileng, A.~Whata, M.~Olusanya, and S.~Mhlongo, ``Application of convnext
  with transfer learning and data augmentation for malaria parasite detection
  in resource-limited settings using microscopic images,'' \emph{PloS one},
  vol.~20, no.~6, p. e0313734, 2025.

\bibitem{amangeldi2025cnn}
A.~Amangeldi, A.~Taigonyrov, M.~H. Jawad, and C.~E. Mbonu, ``Cnn and vit
  efficiency study on tiny imagenet and dermamnist datasets,'' \emph{arXiv
  preprint arXiv:2505.08259}, 2025.

\bibitem{mehta2021mobilevit}
S.~Mehta and M.~Rastegari, ``Mobilevit: light-weight, general-purpose, and
  mobile-friendly vision transformer,'' \emph{arXiv preprint arXiv:2110.02178},
  2021.

\bibitem{wei2023joint}
S.~Wei, T.~Ye, S.~Zhang, Y.~Tang, and J.~Liang, ``Joint token pruning and
  squeezing towards more aggressive compression of vision transformers,'' in
  \emph{Proceedings of the IEEE/CVF conference on computer vision and pattern
  recognition}, 2023, pp. 2092--2101.

\bibitem{hong2025knowledge}
I.~Hong and C.~Choi, ``Knowledge distillation vulnerability of deit through cnn
  adversarial attack,'' \emph{Neural Computing and Applications}, vol.~37,
  no.~12, pp. 7721--7731, 2025.

\bibitem{graham2021levit}
B.~Graham, A.~El-Nouby, H.~Touvron, P.~Stock, A.~Joulin, H.~J{\'e}gou, and
  M.~Douze, ``Levit: a vision transformer in convnet's clothing for faster
  inference,'' in \emph{Proceedings of the IEEE/CVF international conference on
  computer vision}, 2021, pp. 12\,259--12\,269.

\bibitem{wang2023image}
W.~Wang, H.~Bao, L.~Dong, J.~Bjorck, Z.~Peng, Q.~Liu, K.~Aggarwal, O.~K.
  Mohammed, S.~Singhal, S.~Som \emph{et~al.}, ``Image as a foreign language:
  Beit pretraining for vision and vision-language tasks,'' in \emph{Proceedings
  of the IEEE/CVF Conference on Computer Vision and Pattern Recognition}, 2023,
  pp. 19\,175--19\,186.

\bibitem{shabrina2023deep}
N.~H. Shabrina, R.~A. Lika, and S.~Indarti, ``Deep learning models for
  automatic identification of plant-parasitic nematode,'' \emph{Artificial
  Intelligence in Agriculture}, vol.~7, pp. 1--12, 2023.

\bibitem{zhao2022swin}
D.-z. Zhao, X.-k. Wang, T.~Zhao, H.~Li, D.~Xing, H.-t. Gao, F.~Song, G.-h.
  Chen, and C.-x. Li, ``A swin transformer-based model for mosquito species
  identification,'' \emph{Scientific Reports}, vol.~12, no.~1, p. 18664, 2022.

\bibitem{han2024efficient}
Q.~Han, G.~Zhang, J.~Huang, P.~Gao, Z.~Wei, and S.~Lu, ``Efficient mae towards
  large-scale vision transformers,'' in \emph{Proceedings of the IEEE/CVF
  Winter Conference on Applications of Computer Vision}, 2024, pp. 606--615.

\bibitem{li2022efficientformer}
Y.~Li, G.~Yuan, Y.~Wen, J.~Hu, G.~Evangelidis, S.~Tulyakov, Y.~Wang, and
  J.~Ren, ``Efficientformer: Vision transformers at mobilenet speed,''
  \emph{Advances in Neural Information Processing Systems}, vol.~35, pp.
  12\,934--12\,949, 2022.

\bibitem{olson2016tpot}
R.~S. Olson and J.~H. Moore, ``Tpot: A tree-based pipeline optimization tool
  for automating machine learning,'' in \emph{Workshop on automatic machine
  learning}.\hskip 1em plus 0.5em minus 0.4em\relax PMLR, 2016, pp. 66--74.

\bibitem{tang2024autogluon}
Z.~Tang, H.~Fang, S.~Zhou, T.~Yang, Z.~Zhong, T.~Hu, K.~Kirchhoff, and
  G.~Karypis, ``Autogluon-multimodal (automm): Supercharging multimodal automl
  with foundation models,'' \emph{arXiv preprint arXiv:2404.16233}, 2024.

\bibitem{jin2019auto}
H.~Jin, Q.~Song, and X.~Hu, ``Auto-keras: An efficient neural architecture
  search system,'' in \emph{Proceedings of the 25th ACM SIGKDD international
  conference on knowledge discovery \& data mining}, 2019, pp. 1946--1956.

\bibitem{akiba2019optuna}
T.~Akiba, S.~Sano, T.~Yanase, T.~Ohta, and M.~Koyama, ``Optuna: A
  next-generation hyperparameter optimization framework,'' in \emph{Proceedings
  of the 25th ACM SIGKDD international conference on knowledge discovery \&
  data mining}, 2019, pp. 2623--2631.

\bibitem{lindauer2022smac3}
M.~Lindauer, K.~Eggensperger, M.~Feurer, A.~Biedenkapp, D.~Deng, C.~Benjamins,
  T.~Ruhkopf, R.~Sass, and F.~Hutter, ``Smac3: A versatile bayesian
  optimization package for hyperparameter optimization,'' \emph{Journal of
  Machine Learning Research}, vol.~23, no.~54, pp. 1--9, 2022.

\bibitem{pham2018efficient}
H.~Pham, M.~Guan, B.~Zoph, Q.~Le, and J.~Dean, ``Efficient neural architecture
  search via parameters sharing,'' in \emph{International conference on machine
  learning}.\hskip 1em plus 0.5em minus 0.4em\relax PMLR, 2018, pp. 4095--4104.

\bibitem{luo2018neural}
R.~Luo, F.~Tian, T.~Qin, E.~Chen, and T.-Y. Liu, ``Neural architecture
  optimization,'' \emph{Advances in neural information processing systems},
  vol.~31.

\bibitem{cocchi2025llava}
F. Cocchi, N. Moratelli, D. Caffagni, S. Sarto, L. Baraldi, M. Cornia, and R. Cucchiara,
``Llava-more: A comparative study of llms and visual backbones for enhanced visual instruction tuning,''
in \emph{Proc. IEEE/CVF Int. Conf. Comput. Vis.}, 2025, pp. 4278--4288.

\bibitem{chen2024internvl}
Z. Chen, J. Wu, W. Wang, W. Su, G. Chen, S. Xing, M. Zhong, Q. Zhang, X. Zhu, L. Lu, \emph{et al.},
``Internvl: Scaling up vision foundation models and aligning for generic visual-linguistic tasks,''
in \emph{Proc. IEEE/CVF Conf. Comput. Vis. Pattern Recognit.}, 2024, pp. 24185--24198.


\bibitem{wang2024qwen2}
P. Wang, S. Bai, S. Tan, S. Wang, Z. Fan, J. Bai, K. Chen, X. Liu, J. Wang, W. Ge, \emph{et al.},
``Qwen2-vl: Enhancing vision-language model's perception of the world at any resolution,''
\emph{arXiv preprint arXiv:2409.12191}, 2024.

\bibitem{xiao2024florence}
B. Xiao, H. Wu, W. Xu, X. Dai, H. Hu, Y. Lu, M. Zeng, C. Liu, and L. Yuan,
``Florence-2: Advancing a unified representation for a variety of vision tasks,''
in \emph{Proc. IEEE/CVF Conf. Comput. Vis. Pattern Recognit.}, 2024, pp. 4818--4829.

\bibitem{stevens2024bioclip}
S. Stevens, J. Wu, M. J. Thompson, E. G. Campolongo, C. H. Song, D. E. Carlyn, L. Dong, W. M. Dahdul, C. Stewart, T. Berger-Wolf, \emph{et al.},
``Bioclip: A vision foundation model for the tree of life,''
in \emph{Proc. IEEE/CVF Conf. Comput. Vis. Pattern Recognit.}, 2024, pp. 19412--19424.


\bibitem{gu2025bioclip}
J. Gu, S. Stevens, E. G. Campolongo, M. J. Thompson, N. Zhang, J. Wu, A. Kopanev, Z. Mai, A. E. White, J. Balhoff, \emph{et al.},
``Bioclip 2: Emergent properties from scaling hierarchical contrastive learning,''
\emph{arXiv preprint arXiv:2505.23883}, 2025.


\end{thebibliography}
\end{document}